\documentclass{article}

\usepackage[accepted]{icml2024}

\usepackage[colorlinks=true,allcolors=perfblue]{hyperref} 
\usepackage[utf8]{inputenc} %
\usepackage[T1]{fontenc}    %
\usepackage{url}            %
\usepackage{booktabs}       %
\usepackage{amsfonts}       %
\usepackage{nicefrac}       %
\usepackage{microtype}      %
\usepackage{lipsum}		%
\usepackage{graphicx}
\usepackage{natbib}
\usepackage{doi}
\usepackage{amsmath}
\usepackage{algorithm}
\usepackage{algorithmic}
\usepackage{amssymb}
\usepackage{enumitem}
\usepackage{pifont}
\usepackage{tikz}
\usetikzlibrary{arrows,calc,matrix,backgrounds}
\newcommand{\tikzAngleOfLine}{\tikz@AngleOfLine}
\def\tikz@AngleOfLine(#1)(#2)#3{%
\pgfmathanglebetweenpoints{%
\pgfpointanchor{#1}{center}}{%
\pgfpointanchor{#2}{center}}
\pgfmathsetmacro{#3}{\pgfmathresult}%
}

\usepackage[utf8]{inputenc} %
\usepackage[T1]{fontenc}    %
\usepackage{url}            %
\usepackage{booktabs}       %
\usepackage{amsfonts, amsmath, amssymb, bm}       %
\usepackage{nicefrac}       %
\usepackage{microtype}      %
\usepackage{xcolor}         %
\usepackage{amsthm}
\usepackage{thmtools}
\usepackage{graphicx}
\usepackage{float}
\usepackage{algorithm}
\usepackage{algorithmic}
\usepackage{pythonhighlight}

\usepackage{caption}
\usepackage{subcaption}

\DeclareMathOperator*{\argmax}{\arg\!\max}

\usepackage{dsfont}
\usepackage[mode=build]{standalone}

\usepackage[capitalize]{cleveref}

\usepackage{etoc}
\usepackage{cancel}

\usepackage[customcolors]{hf-tikz}
\definecolor{expert}{HTML}{008000}
\definecolor{error}{HTML}{f96565}
\definecolor{learner}{HTML}{F79646}
\definecolor{perfblue}{RGB}{64, 114, 175}

\theoremstyle{plain}
\newtheorem{theorem}{Theorem}[section]

\newtheorem{lemma}[theorem]{Lemma}

\newtheorem{corollary}[theorem]{Corollary}
\theoremstyle{definition}
\newtheorem{definition}[theorem]{Definition}

\theoremstyle{remark}

\declaretheoremstyle[
headfont=\normalfont\itshape,
qed=\qedsymbol,
]{mypf}
\declaretheorem[numbered=no, name=Proof, style=mypf]{pf}

 \renewcommand{\algorithmiccomment}[1]{// #1}
 \usepackage{bbm}
 \usepackage{changepage}

\usepackage{transparent}
\newcommand{\algcommentlight}[1]{\textcolor{perfblue}{\transparent{0.8}\small{\texttt{\textbf{//\hspace{2pt}#1}}}}}

\usepackage{nicematrix}

\icmltitlerunning{Hybrid Inverse Reinforcement Learning}

\begin{document}

\twocolumn[
\icmltitle{Hybrid Inverse Reinforcement Learning}

\icmlsetsymbol{equal}{*}

\begin{icmlauthorlist}
\icmlauthor{Juntao Ren}{equal,yyy}
\icmlauthor{Gokul Swamy}{equal,comp}
\icmlauthor{Zhiwei Steven Wu}{comp}
\icmlauthor{J. Andrew Bagnell}{sch,comp}
\icmlauthor{Sanjiban Choudhury}{yyy}
\end{icmlauthorlist}

\icmlaffiliation{yyy}{Cornell University}
\icmlaffiliation{comp}{Carnegie Mellon University}
\icmlaffiliation{sch}{Aurora Innovation}

\icmlcorrespondingauthor{Gokul Swamy}{gswamy@cmu.edu}

\icmlkeywords{Machine Learning, ICML}

\vskip 0.3in
]

\printAffiliationsAndNotice{\icmlEqualContribution} %

\begin{abstract}
The inverse reinforcement learning approach to imitation learning is a double-edged sword. On the one hand, it can enable learning from a smaller number of expert demonstrations with more robustness to error compounding than behavioral cloning approaches. On the other hand, it requires that the learner repeatedly solve a computationally expensive reinforcement learning (RL) problem. Often, much of this computation is wasted searching over policies very dissimilar to the expert's. In this work, we propose using \textit{hybrid RL} -- training on a mixture of online and expert data -- to curtail unnecessary exploration. Intuitively, the expert data focuses the learner on good states during training, which reduces the amount of exploration required to compute a strong policy. Notably, such an approach doesn't need the ability to reset the learner to arbitrary states in the environment, a requirement of prior work in efficient inverse RL. More formally, we derive a reduction from inverse RL to \textit{expert-competitive RL} (rather than globally optimal RL) that allows us to dramatically reduce interaction during the inner policy search loop while maintaining the benefits of the IRL approach. This allows us to derive both model-free and model-based hybrid inverse RL algorithms with strong policy performance guarantees. Empirically, we find that our approaches are significantly more sample efficient than standard inverse RL and several other baselines on a suite of continuous control tasks.

\end{abstract}

\section{Introduction}

\label{sec:intro}
Broadly speaking, we can break down the approaches to imitation learning (IL) into \textit{offline} algorithms (e.g. behavioral cloning, \citep{pomerleau1988alvinn}) and \textit{interactive} algorithms (e.g. inverse reinforcement learning \citep{MaxEntIRL}, DAgger \citep{ross2011reduction}). 
Offline approaches to imitation aren't robust to the covariate shift between the expert's state distribution and the learner's \textit{induced} state distribution; therefore, they suffer from compounding errors which results in poor test-time performance \citep{ross2011reduction, swamy2021moments, wang2021instabilities}. %
Instead, interactive algorithms allow the the learner to observe the consequences of their actions and therefore learn to recover from their own mistakes. This is the fundamental reason why interactive approaches like inverse reinforcement learning (IRL) continue to provide state-of-the-art performance for challenging tasks like autonomous driving~\citep{bronstein2022hierarchical, igl2022symphony, vinitsky2022nocturne} and underlie large-scale services like Google Maps~\citep{barnes2023massively}.

\begin{figure}
     \centering
     \includegraphics[width=0.45\textwidth]{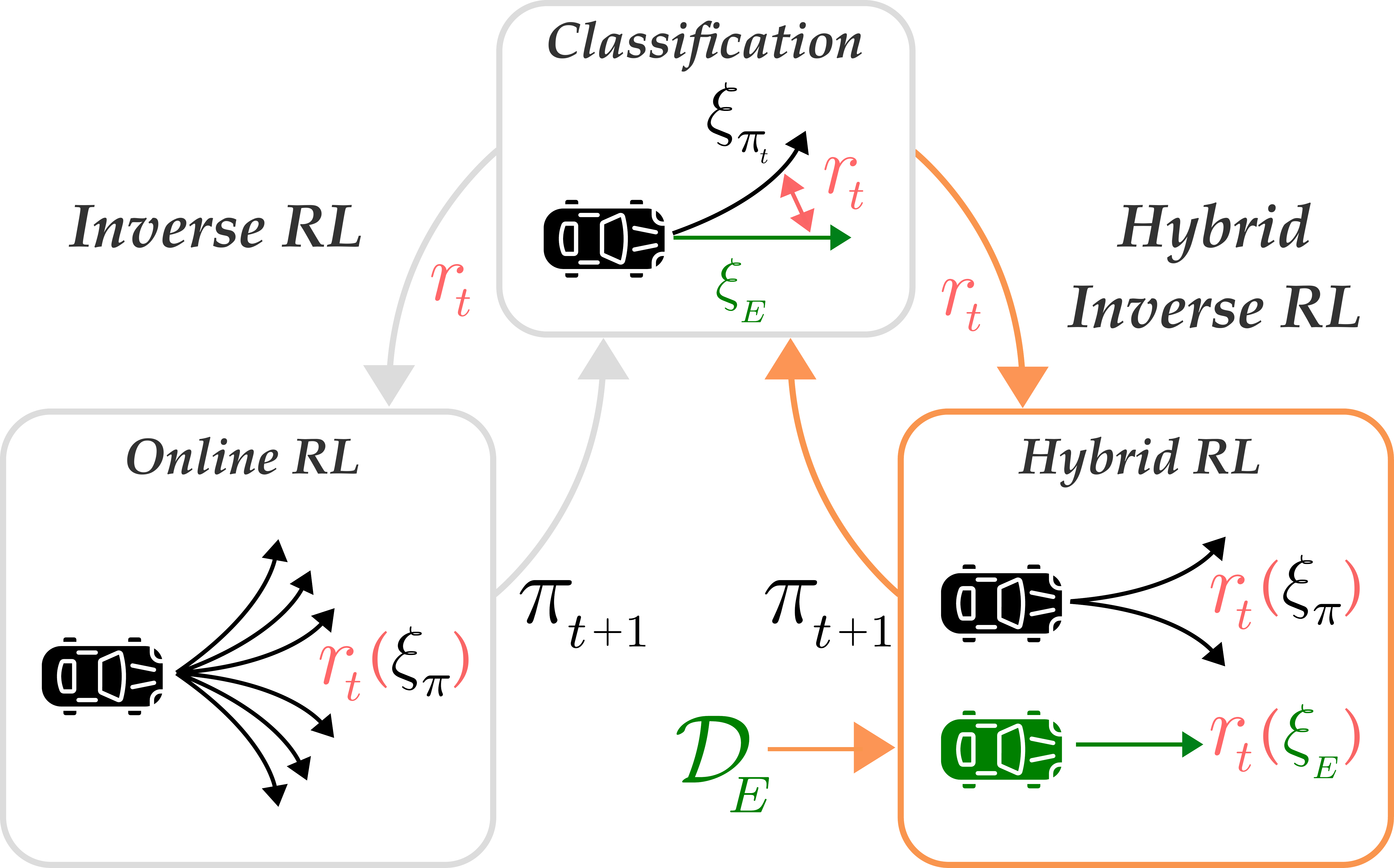}
      \caption{
     Standard inverse reinforcement algorithms (left) require repeatedly solving a reinforcement learning problem in their inner loop. Thus, the learner is potentially forced to explore the entire state space to find any reward. We introduce \textit{hybrid inverse reinforcement learning}, where the learner trains on a mixture of its own and the expert's data during the policy search inner loop. This reduces the exploration burden on the learner by providing positive examples. We provide model-free and model-based algorithms that are both significantly more sample efficient than standard inverse RL approaches on continuous control tasks.}
     \label{fig:ffig}
 \end{figure}

However, inverse reinforcement learning reduces the problem of imitation to repeatedly solving a reinforcement learning problem, and thus the agent has to potentially pay the exponential interaction complexity of reinforcement learning \citep{kakade2003sample}. When performed in the real world, this interaction can be both unsafe and time-consuming; in simulation, it incurs great computational expense. This fact motivates our key question: \textit{how can we reduce the amount of interaction performed in inverse reinforcement learning?}

At its core, the reason reinforcement learning is interaction-inefficient is because of global \textit{exploration}: in the worst case, the learner needs to reach all possible states to figure out what decisions are optimal over the horizon.
This means that in IRL, the learner often spends the majority of interactions trying out policies that are quite dissimilar to the expert's in the hope of finding a bit of reward. This is rather odd, given our goal is just to imitate the expert. Put differently, when optimizing a potential reward function, we should only be competing against policies with similar visitation distributions to the expert. We therefore narrow our overarching question: \textit{how do we focus IRL policy search on policies that are similar to the expert's?} %

Recent work by \citet{Swamy2023InverseRL} shows that one can dramatically reduce the amount of exploration required by resetting the learner to states from the expert demonstrations during policy search. 
Despite this approach's strong theoretical guarantees, the ability to reset the learner to arbitrary states has limited feasibility in the real world. Thus, we focus on how we can curtail unnecessary exploration \textit{without} assuming \textit{generative model} access to the environment. 

We provide a general reduction that allows one to use \textit{any} RL algorithm that merely guarantees returning a policy that competes with the expert (rather than competing with the optimal policy) for policy search in IRL. This generalizes the argument of \citet{Swamy2023InverseRL}, allowing us to leverage efficient RL algorithms that don't require resets for IRL.

Specifically, we propose to use \textit{hybrid} reinforcement learning \citep{ross2012agnostic, song2022hybrid, zhou2023offline} to speed up the policy search component of inverse reinforcement learning. 
In hybrid RL, one trains a policy to do well on \textit{both} the offline data and the distribution of data it induces (e.g. by using data from both buffers when fitting a $Q$-function). Rather than competing against an arbitrary policy (as we do in online RL which therefore leads to having to pay for extensive exploration), this procedure asks only the learner to compete against policies \textit{covered} by the offline dataset. Hybrid RL gives similar theoretical guarantees as offline RL 
without requiring explicit pessimism (which can be brittle in practice and intractable in theory) and retains a higher degree of robustness to covariate shift. Given our goal is to compete with the expert, we propose to simply use the expert demonstrations as the offline dataset for hybrid RL. Our key insight is that \textbf{\textit{we can use hybrid RL as the policy search procedure in IRL to curtail exploration.}}

More explicitly, the contributions of our work are three-fold:

\textbf{1. We provide a reduction from inverse RL to expert-competitive RL}. We prove that as long as our policy search procedure guarantees to output a sequence of policies that competes with the expert \textit{on average} over a sequence of chosen rewards, we are able to compute a policy that competes with the expert on the ground truth reward. Notably, our reduction generalizes the underlying argument of \citet{Swamy2023InverseRL} to methods that don't require the ability to reset the learner to arbitrary states in the environment.

\textbf{2. We derive two hybrid inverse RL algorithms: model-free \texttt{HyPE} and model-based \texttt{HyPER}}. \texttt{HyPE} uses the HyQ algorithm of \citet{song2022hybrid} in its inner loop while \texttt{HyPER} uses the LAMPS algorithm of \citet{vemula2023virtues} as its policy search procedure. We provide performance guarantees for both algorithms and discuss their pros and cons relative to other fast inverse RL algorithms.

\textbf{3. We demonstrate that on a suite of continuous control tasks, \texttt{HyPE} and \texttt{HyPER} are significantly more sample efficient than standard approaches}. In addition to out-performing GAIL-like approaches \citep{ho2016generative} and behavioral cloning, we also find that we are able to consistently out-perform the FILTER algorithm of \cite{Swamy2023InverseRL} and IQLearn \citep{garg2021iqlearn}. %

\section{Related Work}

\noindent \textbf{Hybrid Reinforcement Learning.} Hybrid RL --- using data that covers a strong policy %
to speed up reinforcement learning --- comes in several variants. One variant is to \textit{reset} the learner to states from the offline distribution \citep{Bagnell2003PolicySB, Ross2014ReinforcementAI}. Another is \textit{hybrid training}: using a combination of on-policy data from the learner and the offline distribution when fitting the critic \citep{song2022hybrid, zhou2023offline} or the model \citep{ross2012agnostic, vemula2023virtues} used in the policy update. While the former variant comes with stronger guarantees as far as interaction complexity, the latter is more applicable to a wider variety of problems due to weaker reset requirements~\citep{Hester2018DeepQF, ball2023efficient, luo2023serl}. 
We lift the insights from two hybrid training algorithms --- the HyQ algorithm of \citet{song2022hybrid} and the LAMPS algorithm of \citep{vemula2023virtues} --- to the space of imitation learning.

\noindent \textbf{Sample-Efficient Inverse Reinforcement Learning.}
Various lines of work have attempted to address the sample-inefficiency of the RL inner loop of inverse RL. One line removes the inner loop entirely by using the difference of $Q$ functions across timesteps as an implicit reward \citep{Dvijotham2010InverseOC, garg2021iqlearn}. However, since $Q$ functions depend on the dynamics of the environment, it is unclear if such methods will produce consistent estimates of the expert's policy if data was collected across agents with slightly different dynamics (e.g. cars with different wheel frictions).
On the other hand, reward-based methods like ours have repeatedly demonstrated robustness to this issue \citep{silver2010learning, ratliff2009learning, kolter2008control, ng2006autonomous, zucker2011optimization, ziebart2008navigate}.

Another line of work attempts to use resets to the expert's state distribution to curtail the exploration the learner performs during the inner loop \citep{Swamy2023InverseRL}. This approach comes with strong guarantees for interaction efficiency but requires generative model access to the environment. Our approach operates under weaker reset assumptions but can only provide weaker guarantees. More explicitly, we lift the reset flavor of hybrid RL to imitation, while we lift hybrid training. We provide a more in-depth comparison efficient IRL methods in Section \ref{sec:battle}.

\noindent \textbf{Hybrid Training for Inverse Reinforcement Learning.}
Perhaps the most similar approaches to our model-free \texttt{HyPE} algorithm are the SQIL approach of \citet{Reddy2019SQILIL} and the AdRIL approach of \citet{swamy2021moments}. Both approaches use data from both the learner and the expert during policy updates by sampling from two separate replay buffers. However, neither of these works rigorously addresses the effect of using off-policy data in the policy optimization subroutine of inverse RL. We provide a general reduction that specifies the properties required for doing so while preserving performance guarantees. In concurrent work, \citet{kolev2024efficient} combine hybrid training with pessimism for more interaction-efficient model-based IRL, using a model ensemble disagreement auxiliary cost in their practical RL procedure. In contrast, we focus on the benefits hybrid training + expert resets provides for model-based IRL. This allows us to elide intractable pessimism in theory and ensemble-based approximations in practice.

\section{Hybrid Inverse RL}
\begin{algorithm}[t]
\begin{algorithmic}
\STATE {\bfseries Input:} Demos. $\mathcal{D}_E$, Policy class $\Pi$, Reward class $\mathcal{F}_r$
\STATE {\bfseries Output:} Trained policy $\pi$
\STATE Initialize $\pi_1 \in \Pi$
\FOR{$t$ in $1 \dots T$}
\STATE \algcommentlight{Use any no-regret algo to pick $f$}
\STATE $f_{t} \gets f_{t - 1} + \nabla_f (J(\pi_E, f) - J(\pi_{t}, f))$
\STATE $\pi_{t+1} \gets \texttt{RL}(r=f_t, \Pi=\Pi)$

\ENDFOR
\STATE {\bfseries Return } mixture of $\pi_{1:T}$. 
\end{algorithmic}
\caption{\texttt{(Dual) IRL} ( \citet{ziebart2008maximum}) \label{alg:dual-irl}}
\end{algorithm}
\subsection{A Game-Theoretic Perspective on Inverse RL}
We consider a finite-horizon Markov Decision Process (MDP) \citep{puterman2014markov} parameterized by $\langle \mathcal{S}, \mathcal{A}, \mathcal{T}, r, H \rangle$ where $\mathcal{S}$, $\mathcal{A}$ are the state and action spaces, $\mathcal{T}: \mathcal{S} \times \mathcal{A}\rightarrow \Delta(\mathcal{S})$ is the transition operator, $r: \mathcal{S} \times \mathcal{A} \rightarrow [-1, 1]$ is the reward function, and $H$ is the horizon. In the inverse RL setup, we see trajectories generated by an expert policy $\pi_E: \mathcal{S} \rightarrow \Delta(\mathcal{A})$, but do not know the reward function. Our goal is to find a policy that, no matter what reward function we are evaluated under, performs as well as the expert. We cast this problem as a zero-sum game between a policy player and an adversary that tries to pick out differences between expert and learner policies \citep{syed2007game, swamy2021moments}. More formally, we optimize over policies $\pi: \mathcal{S} \rightarrow \Delta(\mathcal{A}) \in \Pi$ and reward functions $f: \mathcal{S} \times \mathcal{A} \rightarrow [-1, 1] \in \mathcal{F}_r$. We use $\xi = (s_1,a_1,r_1,\ldots, s_H,a_H,r_H)$ to denote the trajectory generated by some policy. For theoretical simplicity, we assume that our strategy spaces ($\Pi$ and $\mathcal{F}_r$) are convex and compact, that $\mathcal{F}_r$ is closed under negation, and that $r \in \mathcal{F}_r, \pi_E \in \Pi$. Using $J(\pi, \hat{r}) = \mathbb{E}_{\xi \sim \pi}[\sum_{h=1}^H \hat{r}(s_h, a_h)]$  to denote the value of policy $\pi$ under reward function $\hat{r}$, we can express our objective as
\begin{equation}
\min_{\pi \in \Pi} \max_{f \in \mathcal{F}_r} J(\pi_E, f) - J(\pi, f).
\end{equation}

Perhaps the most common strategy for solving this game is to use a no-regret strategy for reward selection against a best-response strategy for policy selection \citep{ziebart2008maximum, swamy2021moments}, which we outline in Algorithm \ref{alg:dual-irl}.
Explicitly, in the \textit{dual} flavor of IRL, one performs a \textit{best-response} via RL at each inner iteration by computing the optimal policy for the adversarially chosen reward function. 

\begin{figure}[h]
    \centering
    \includegraphics[scale=0.32]{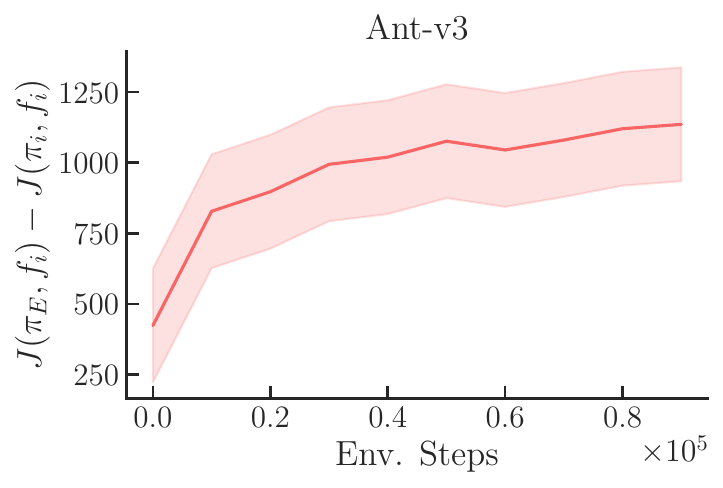}
    \includegraphics[scale=0.32]{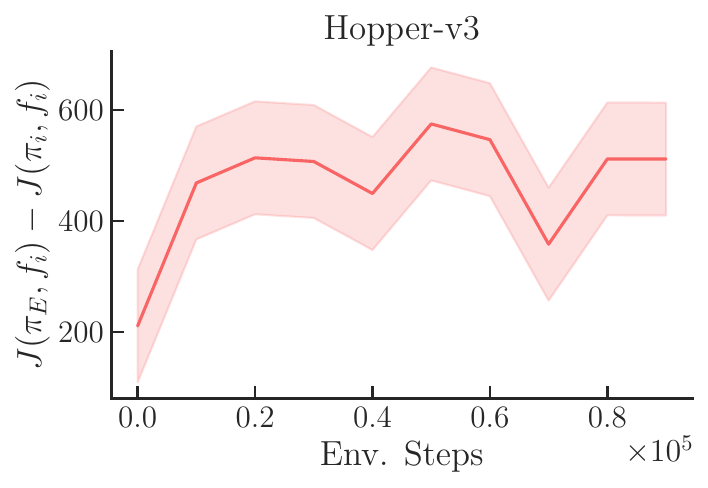} \\
    \includegraphics[scale=0.32]{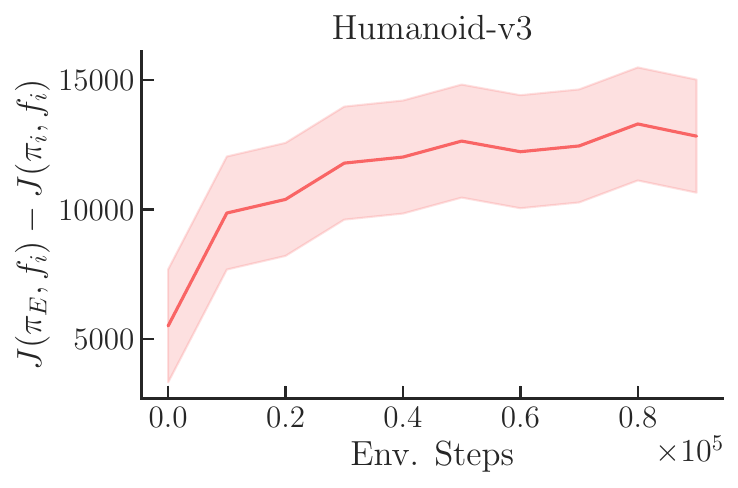}
    \includegraphics[scale=0.32]{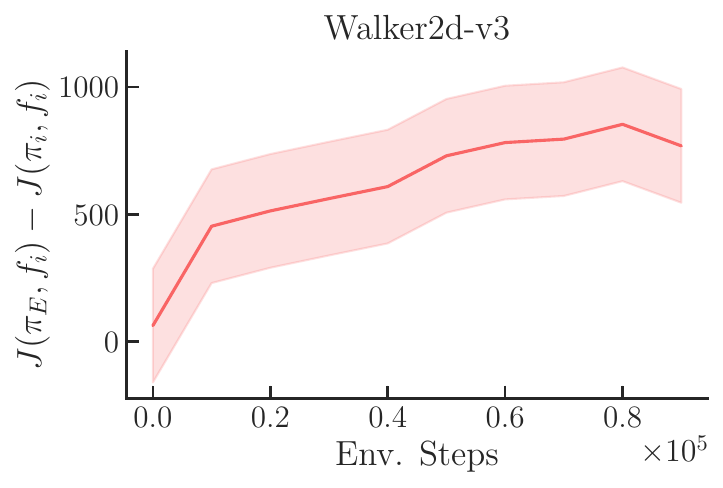}
    \caption{Difference in rewards between the learner policy $\pi_i$ and expert policy $\pi_E$ under the discriminator function $f_i$ for the first 100k environment interactions in primal IRL.}
    \label{fig:reward-diff}
\end{figure}

Consider the learner operating in a tree-structured MDP and let $\mathcal{F}_r$ be the class of reward functions that are 0 everywhere except for a single leaf node. Then, to find any reward, the learner needs to explore the entire tree at each iteration, an amount of interaction that scales exponentially with the task horizon. %
This construction is more than a pathological example: Fig.~\ref{fig:reward-diff} shows that even for \textit{primal} approaches (e.g. GAIL) that don't optimize the reward to completion in their inner loop (and instead perform a small no-regret update -- i.e. a gradient step), we observe empirical evidence of the adversary being able to consistently pick out differences between the expert and the learner, and drive up the value difference $J(\pi_E, f_i) - J(\pi, f_i)$. Even as the learner slowly improves upon the current reward, the adversary repeatedly shifts the reward function, thus driving up interactions and introducing instability. Effectively, any adversarial IRL algorithm (whether primal, dual or a mixture) ends up solving a potentially hard global exploration problem \textit{at least} once.

Indeed, one may need to explore the entire state space for the optimal policy when given an arbitrary reward function. However, in inverse RL we explicitly choose reward functions that rate the expert higher than the history of learner policies. Thus, the fact that the standard recipe for both primal and dual IRL completely ignore the expert demonstrations during policy optimization seems suboptimal. This begs the question: \textit{how can we give the learner examples of expert behavior during policy optimization to reduce the amount of exploration required to find good states?} We now discuss the core principle that underlies multiple ways to do so while preserving performance guarantees.

\subsection{Expert-Relative Regret Oracles in Inverse RL}
\begin{figure}[h]
    \begin{subfigure}[b]{0.23\textwidth}
        \centering
        $f_1$\\[3pt]
            \centering
    \begin{tikzpicture}[scale=1, transform shape]
    \node (1) [draw, very thick, circle, minimum size=0.7cm, color=expert] at (0, 0) {\scriptsize $0$};
    \node (2)  [draw, very thick, circle, minimum size=0.7cm, color=expert]  at (-1, -.75) {\scriptsize $0$}; 
    \node (3) [draw, very thick, circle, minimum size=0.7cm, color=learner]  at (1, -.75) {\scriptsize $0$};
    
    \node (4) [draw, very thick, circle, minimum size=0.7cm, color=expert]  at (-1.5, -1.75) {\scriptsize $+1$};
    \node (5)  [draw, very thick, circle, minimum size=0.7cm, color=error]  at (-0.5, -1.75) {\scriptsize $0$};
    \node (6) [draw, very thick, circle, minimum size=0.7cm, color=error]  at (0.5, -1.75) {\scriptsize $0$};
    \node (7) [draw, very thick, circle, minimum size=0.7cm, color=learner]  at (1.5, -1.75) {\scriptsize $+1$};

    \draw [->, very thick, color=expert] (1) to (2);
    \draw [->, very thick, color=learner] (1) to (3);
    
    \draw [->, very thick, color=expert] (2) to (4);
    \draw [->, very thick, color=error] (2) to (5);
    
    \draw [->, very thick, color=error] (3) to (6);
    \draw [->, very thick, color=learner] (3) to (7);

    \node [circle, minimum size=0.5cm](a) at ([{shift=(350:0.6)}]7){};
    \coordinate (b) at (intersection 1 of 7 and a);
    \coordinate (c) at (intersection 2 of 7 and a);
    \tikzAngleOfLine(a)(b){\AngleStart}
    \tikzAngleOfLine(a)(c){\AngleEnd}
    \draw[very thick,->, color=learner]%
    let \p1 = ($ (a) - (b) $), \n2 = {veclen(\x1,\y1)}
    in
        (b) arc (\AngleStart + 200:\AngleEnd+45:\n2);

    \node [circle, minimum size=0.5cm](d) at ([{shift=(350:0.6)}]6){};
    \coordinate (e) at (intersection 1 of 6 and d);
    \coordinate (f) at (intersection 2 of 6 and d);
    \tikzAngleOfLine(d)(e){\AngleStart}
    \tikzAngleOfLine(d)(f){\AngleEnd}
    \draw[very thick,->, color=error]%
    let \p1 = ($ (d) - (e) $), \n2 = {veclen(\x1,\y1)}
    in
        (e) arc (\AngleStart + 200:\AngleEnd+45:\n2);

    \node [circle, minimum size=0.5cm](g) at ([{shift=(350:0.6)}]5){};
    \coordinate (h) at (intersection 1 of 5 and g);
    \coordinate (i) at (intersection 2 of 5 and g);
    \tikzAngleOfLine(g)(h){\AngleStart}
    \tikzAngleOfLine(g)(i){\AngleEnd}
    \draw[very thick,->, color=error]%
    let \p1 = ($ (g) - (h) $), \n2 = {veclen(\x1,\y1)}
    in
        (h) arc (\AngleStart + 200:\AngleEnd+45:\n2);

    \end{tikzpicture} %
        \footnotesize
        {${\color{learner}\pi_1} = \argmax_{\pi \in \Pi} J(\pi, f_1)$}
    \end{subfigure}
    \begin{subfigure}[b]{0.23\textwidth}
        \centering 
        $f_2$\\[3pt]
            \begin{tikzpicture}[scale=1, transform shape]
    \node (1) [draw, very thick, circle, minimum size=0.7cm, color=expert] at (0, 0) {\scriptsize $0$};
    \node (2)  [draw, very thick, circle, minimum size=0.7cm, color=expert]  at (-1, -.75) {\scriptsize $0$}; 
    \node (3) [draw, very thick, circle, minimum size=0.7cm, color=learner]  at (1, -.75) {\scriptsize $0$};
    
    \node (4) [draw, very thick, circle, minimum size=0.7cm, color=expert]  at (-1.5, -1.75) {\scriptsize $+1$};
    \node (5)  [draw, very thick, circle, minimum size=0.7cm, color=error]  at (-0.5, -1.75) {\scriptsize $0$};
    \node (6) [draw, very thick, circle, minimum size=0.7cm, color=learner]  at (0.5, -1.75) {\scriptsize $+1$};
    \node (7) [draw, very thick, circle, minimum size=0.7cm, color=error]  at (1.5, -1.75) {\scriptsize $0$};

    \draw [->, very thick, color=expert] (1) to (2);
    \draw [->, very thick, color=learner] (1) to (3);
    
    \draw [->, very thick, color=expert] (2) to (4);
    \draw [->, very thick, color=error] (2) to (5);
    
    \draw [->, very thick, color=learner] (3) to (6);
    \draw [->, very thick, color=error] (3) to (7);

    \node [circle, minimum size=0.5cm](a) at ([{shift=(350:0.6)}]7){};
    \coordinate (b) at (intersection 1 of 7 and a);
    \coordinate (c) at (intersection 2 of 7 and a);
    \tikzAngleOfLine(a)(b){\AngleStart}
    \tikzAngleOfLine(a)(c){\AngleEnd}
    \draw[very thick,->, color=error]%
    let \p1 = ($ (a) - (b) $), \n2 = {veclen(\x1,\y1)}
    in
        (b) arc (\AngleStart + 200:\AngleEnd + 45:\n2);

    \node [circle, minimum size=0.5cm](d) at ([{shift=(350:0.6)}]6){};
    \coordinate (e) at (intersection 1 of 6 and d);
    \coordinate (f) at (intersection 2 of 6 and d);
    \tikzAngleOfLine(d)(e){\AngleStart}
    \tikzAngleOfLine(d)(f){\AngleEnd}
    \draw[very thick,->, color=learner]%
    let \p1 = ($ (d) - (e) $), \n2 = {veclen(\x1,\y1)}
    in
        (e) arc (\AngleStart + 200:\AngleEnd+45:\n2);

    \node [circle, minimum size=0.5cm](g) at ([{shift=(350:0.6)}]5){};
    \coordinate (h) at (intersection 1 of 5 and g);
    \coordinate (i) at (intersection 2 of 5 and g);
    \tikzAngleOfLine(g)(h){\AngleStart}
    \tikzAngleOfLine(g)(i){\AngleEnd}
    \draw[very thick,->, color=error]%
    let \p1 = ($ (g) - (h) $), \n2 = {veclen(\x1,\y1)}
    in
        (h) arc (\AngleStart + 200:\AngleEnd+45:\n2);

    \end{tikzpicture} %
        \footnotesize
        {${\color{learner}\pi_2} = \argmax_{\pi \in \Pi} J(\pi, f_2)$}
    \end{subfigure}
    \caption{\label{fig:tree_mdp} Consider a binary tree MDP. Define $\Pi$ to be the set of all deterministic policies (paths through the tree), and $\mathcal{F}_r$ the class of rewards that always assign $+1$ to the bottom-left node and an additional $+1$ to any one of the three other leaf nodes. The expert (the green path) always takes the leftmost path. Note that the expert is not optimal under any $f \in \mathcal{F}_r$. In the first image, the learner (the orange path) has computed the best response to $f_1$ (the labels on the nodes). To penalize the learner, $f_2$ shifts the reward to a neighboring leaf node. As a result, $\pi_2$ must search through the entire tree to compute the best-response. Beyond the repeated exploration required to compute a best-response, the best responses are different across iterations, which leads to instability in policy training.}
\end{figure}
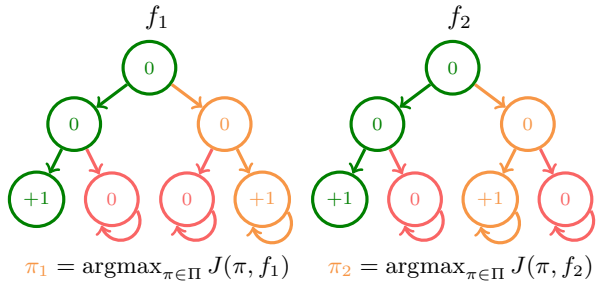

At heart, the game-theoretic approach to inverse reinforcement learning relies on the following intuition: \textit{we must have found a policy with performance close to the expert's if there is no reward function that can tell the difference between the learner's current policy and the expert's.}  We often operationalize this principle by repeatedly picking reward functions that score the expert higher than the learner and then computing the optimal policy under this reward. Critically, however, there is no reason for the expert to be the \textit{optimal} policy under a proposed reward function -- it merely needs to score higher than the learner. We provide a simple example of this point in Fig.~\ref{fig:tree_mdp}.
 
Thus, when computing a \textit{best response} to this reward function (i.e. the optimal policy), the learner often needs to try out a variety of policies that visit states quite different to those the expert visits. This has two negative consequences: first, it may necessitate a large amount of interaction per iteration. Second, because the optimal policy can vary wildly across iterations, it can introduce instability into the training process. Both concerns still apply even for primal approaches like GAIL \citep{ho2016generative} that take a small no-regret step at each iteration.\footnote{To see why the former point applies, observe that even if we were to restrict the adversary to only picking the ground truth $r$, a primal approach would eventually involve computing the optimal policy, which requires extensive exploration in the worst case.} If we pause and take a step back, this is somewhat odd: given our goal is merely to compete with the expert under a variety of potential reward functions, trying to move towards the \textit{optimal policy} under an adversarially chosen metric seems needlessly expensive. Instead, one might hope that as long as the learner can consistently compete with the \textit{expert} under whatever metric the adversary chooses, we should be able to guarantee that we compete with the expert under the ground-truth reward.

We can formalize our preceding intuition via the notion of an \textit{\textbf{E}xpert-\textbf{R}elative \textbf{R}egret \textbf{Or}acle} ($\mathsf{ERROr}$).
\begin{definition}[$\mathsf{ERROr}\{\mathsf{Reg}_{\pi}(T)\}$] A policy-selection algorithm $\mathbb{A}_{\pi}$ satisfies the $\mathsf{Reg}_{\pi}(T)$ expert-relative regret guarantee if given any sequence of reward functions $f_{1:T}$, it produces a sequence of policies $\pi_{t+1} = \mathbb{A}_{\pi}(f_{1:t})$ such that
\begin{equation}
\sum_{t=1}^T J(\pi_E, f_t) - J(\pi_t, f_t) \leq \mathsf{Reg}_{\pi}(T).
\end{equation}
\label{def:error}
\end{definition}
Critically, this definition does not require us to compute or even compete against the optimal policy for each $f_t$. \footnote{Note that this is a weaker requirement than the per-iteration guarantee the proofs of \citet{Swamy2023InverseRL} require. While this distinction may seem unimportant prima facie, the model-based hybrid RL algorithms we build only guarantee competing with the expert on average \citep{ross2012agnostic, vemula2023virtues}.}

We also define a no-regret reward-selection algorithm. %
\begin{definition} Define $\ell_t(f) = \frac{1}{H} (J(\pi_t, f) - J(\pi_E, f))$. $\mathbb{A}_f$ is a \textit{no-regret reward selection algorithm} if when given a sequence of loss functions  $\ell_{1:t}$ induced by a sequence of policies $\pi_{1:t}$, it produces iterates $f_{t+1} = \mathbb{A}_f(\ell_{1:t})$ such that
\begin{equation}
    \sum_{t=1}^T \ell_t(f_t) - \min_{f^{\star} \in \mathcal{F}_r} \sum_{t=1}^T \ell_t(f^{\star}) \leq \mathsf{Reg}_f(T),
\end{equation}
with $\lim_{T \to \infty} \frac{\mathsf{Reg}_f(T)}{T} = 0.$ 
\label{def:rewsel}
\end{definition}
Due to the linearity of $\ell_t$, standard online no-regret algorithms like gradient descent satisfy this above condition \citep{zinkevich2003online}. We now provide a simple proof that the combination of the above two oracles allows us to efficiently compute a policy with similar performance to the expert's.

\begin{theorem} Assume access to an $\mathbb{A}_{\pi}$ and $\mathbb{A}_f$ that satisfy Definitions \ref{def:error} and \ref{def:rewsel} respectively. Set $\pi_{t+1} = \mathbb{A}_{\pi}(f_{1:t})$ and $f_{t+1} = \mathbb{A}_f(\ell_{1:t})$. Then, $\bar{\pi}$ (the mixture of $\pi_{1:T}$) satisfies
\begin{equation}
J(\pi_E, r) - J(\Bar{\pi}, r) \leq \frac{\mathsf{Reg}_{\pi}(T)}{T} + \frac{\mathsf{Reg}_f(T)}{T} H.
\end{equation}
\label{thm:hyirl}
\end{theorem}
\begin{pf}
\begin{align*}
J(\pi_E, r) - J(\Bar{\pi}, r) &= \frac{1}{T} \sum_{t=1}^T J(\pi_E, r) - J(\pi_t, r) \\
 &\leq \max_{f^{\star} \in \mathcal{F}_r} \frac{1}{T} \sum_{t=1}^T J(\pi_E, f^{\star}) - J(\pi_t, f^{\star}) \\
 &\leq  \frac{1}{T} \sum_{t=1}^T J(\pi_E, f_t) - J(\pi_t, f_t) \\ &+ \frac{\mathsf{Reg}_f(T)}{T} H \\
 &\leq  \frac{\mathsf{Reg}_{\pi}(T)}{T} + \frac{\mathsf{Reg}_f(T)}{T} H.
\end{align*}
\end{pf}
As $T \to \infty$, the second term in the above bound goes to 0 due to the no-regret property of $\mathbb{A}_f$. Thus, in the limit, we only pay for our average policy optimization error relative to the expert. More explicitly, the above result implies that rather than the per-iteration best response we require in dual IRL algorithms like MaxEnt IRL \citep{ziebart2008maximum},
\begin{equation}
\pi_t = \argmax_{\pi \in \Pi} J(\pi, f_t),
\end{equation}
or the no-regret property required to prove guarantees for primal IRL algorithms like GAIL \citep{ho2016generative},
\begin{equation}
\lim_{T \to \infty} \max_{\pi^{\star} \in \Pi}\frac{1}{T }\sum_{t=1}^T J(\pi_t, f_t) - J(\pi^{\star}, f_t) = 0,
\end{equation}
we instead merely need to compete with the expert on average to ensure that we learn a policy with strong performance.

Our above discussion begs the question of whether there are efficient algorithms that satisfy the $\mathsf{ERROr}$ property. Two algorithms that do are the PSDP algorithm of \citet{Bagnell2003PolicySB} and the NRPI algorithm of \citet{Ross2014ReinforcementAI}. \citet{Swamy2023InverseRL} essentially use this property in the proofs of their MMDP and NRMM algorithms. Unfortunately, both PSDP and NRPI require the ability to reset the learner to arbitrary states in the environment, which means MMDP and NRMM do as well. Thus we ask the question: \textit{are there algorithms that satisfy the $\mathsf{ERROr}$ property without requiring generative model access to the environment?} As we detail in the following sections, hybrid RL algorithms answer this question in the affirmative, boding well for their application to the imitation learning setting.

\subsection{\texttt{HyPE}: Model-Free Hybrid Inverse RL}
We now consider how to construct a \textit{model-free} hybrid IRL algorithm. We begin by considering the forward problem. Many model-free hybrid RL algorithms follow the rough structure we outline in Algorithm \ref{alg:hyrl}: use a mixture of on-policy data from the learner and off-policy data from the expert to fit a $Q$ function that we then use for a policy update. For example, the HyQ algorithm of \citet{song2022hybrid} runs Fitted Q Iteration for the critic update and then returns the greedy policy from that critic update for the actor update. The HNPG / HAC algorithms of \citet{zhou2023offline} performs a similar critic update before using the Natural Policy Gradient algorithm of \citet{kakade2001natural} / soft policy iteration algorithm of \citet{ziebart2008maximum} as the actor update. In practice, it is common to just run an off-policy RL algorithms like Soft Actor Critic \citep{Haarnoja2018SoftAO} with expert data in the replay buffer \citep{ball2023efficient}. For the purposes of analysis however, we assume the learner picks policies by running HyQ with the expert demonstrations $\mathcal{D}_E$ as the offline dataset and $-f_t$ as the reward function. As argued by \citet{song2022hybrid}, under certain assumptions (e.g. Bellman Completeness of $\mathcal{F}_Q$ and an MDP with low Bilinear Rank \citep{du2021bilinear}), running HyQ for $M$ steps guarantees that the average of the $M$ policies $\bar{\pi}_{t}$ satisfies
\begin{equation}
J(\pi_E, f_t) - J(\bar{\pi}_{t}, f_t) \leq H^2 O\left(\frac{1}{\sqrt{M}}\right). 
\end{equation}
Observe that if we were to use HyQ to select each $\pi_t$, we would satisfy the $\mathsf{ERRoR}$ property with $\mathsf{Reg}_{\pi}(T) \leq T H^2 O\left(\frac{1}{\sqrt{M}}\right)$. Critically, we do not need the ability to reset the learner to arbitrary states in the environment to run hybrid training algorithms like HyQ, allowing us to curtail exploration \textit{without} generative model access. We refer to this approach as \texttt{HyPE}: \textbf{Hy}brid \textbf{P}olicy \textbf{E}mulation and outline it in Algorithm~\ref{alg:hype}. Running \texttt{HyPE} with HyQ as hybrid RL oracle results in the following performance bound.
\begin{corollary}[\texttt{HyPE} Performance Bound] Consider running \texttt{HyPE} (Algorithm \ref{alg:hype}) with $M$ iterations of HyQ as the hybrid RL subroutine. Then, we have the following:
\begin{equation}
J(\pi_E, r) - J(\Bar{\pi}, r) \leq H^2 O\left(\frac{1}{\sqrt{M}}\right) +  \frac{\mathsf{Reg}_f(T)}{T} H.
\end{equation}
\end{corollary}
\begin{pf}
This follows directly from Theorem \ref{thm:hyirl} and the fact that $\mathsf{Reg}_{\pi}(T) \leq T H^2 O\left(\frac{1}{\sqrt{M}}\right)$, which follows from the fact that the per-iteration sub-optimality with respect to the expert is upper bounded by $H^2 O\left(\frac{1}{\sqrt{M}}\right)$. 
\end{pf}

\begin{algorithm}[t]
\begin{algorithmic}
\STATE {\bfseries Input:} Expert demonstrations $\mathcal{D}_E$, Policy class $\Pi$, Reward class $\mathcal{F}_r$, Critic class $\mathcal{F}_Q$, Learning rate $\eta$.
\STATE {\bfseries Output:} Trained policy $\pi$
\STATE Initialize $f_1 \in \mathcal{F}_r, \pi_1 \in \Pi, Q_1 \in \mathcal{F}_Q$
\FOR{$t$ in $1 \dots T$}
\STATE \algcommentlight{No-regret step over rewards} 
\STATE $f_{t+1} \gets f_t + \eta \nabla_f (J(\pi_t, f) - J(\pi_{E}, f))$
\STATE \algcommentlight{Update policy via hybrid RL}
\STATE $\pi_{t+1}, Q_{t+1} \gets \texttt{HyRL}(\pi_t, Q_t, f_{t+1}, \mathcal{D}_E, \dots)$
\ENDFOR
\STATE {\bfseries Return } best of $\pi_{1:T}$ on validation data.
\end{algorithmic}
\caption{\textbf{Hy}brid \textbf{P}olicy \textbf{E}mulation (\texttt{HyPE}) \label{alg:hype}}
\end{algorithm}

\begin{algorithm}[t]
\begin{algorithmic}
\STATE {\bfseries Input:} Expert demonstrations $\mathcal{D}_E$, Policy class $\Pi$, Critic class $\mathcal{F}_Q$, Batch size $B$, Inner steps $N$, Current policy $\pi_t$, Current critic $Q_t$, Current cost $f_t$.
\STATE {\bfseries Output:} Trained policy $\pi$
\STATE Initialize $\pi_1 = \pi_t, Q_1 = Q_t, \mathcal{D}_{mix} = \{\}$
\FOR{$i$ in $1 \dots N$}
\STATE \algcommentlight{Collect on-policy data}
\STATE $\mathcal{D}_i \gets \{\xi_{1:B} \sim \pi_i\}$
\STATE $\mathcal{D}_{mix} \gets \mathcal{D}_{mix} \cup \mathcal{D}_i \cup \{\xi_{1:B} \sim \mathcal{D}_E\}$
\STATE \algcommentlight{Perform hybrid updates}
\STATE $Q_{i+1} \gets \mathsf{critic\_update}(Q_i, \pi_i, -f_t, \mathcal{D}_{mix}, \mathcal{F}_Q)$
\STATE $\pi_{i+1} \gets \mathsf{actor\_update}(\pi_i, Q_{i+1}, -f_t, \mathcal{D}_{mix}, \Pi)$

\ENDFOR
\STATE {\bfseries Return } Best of $\pi_{1:N}, Q_{1:N}$ on validation data.
\end{algorithmic}
\caption{\textbf{Hy}brid \textbf{RL} (\texttt{HyRL}) \label{alg:hyrl}}
\end{algorithm}

\begin{algorithm}[t]
\begin{algorithmic}
\STATE {\bfseries Input:} Expert demos. $\mathcal{D}_E$, Policy class $\Pi$, Reward class $\mathcal{F}_r$, Batch size $B$, Model class $\mathcal{F}_M$, Learning rate $\eta$.
\STATE {\bfseries Output:} Trained policy $\pi$
\STATE Initialize $f_1 \in \mathcal{F}_r, \pi_1 \in \Pi, M_1 \in \mathcal{F}_M$
\FOR{$t$ in $1 \dots T$}
\STATE \algcommentlight{No-regret step over rewards} 
\STATE $f_{t+1} \gets f_t + \eta \nabla_f \left(J(\pi_t, f) - J(\pi_{E}, f)\right)$
\STATE \algcommentlight{No-regret hybrid step over models}
\STATE $\mathcal{D}_t \gets \{\xi_{1:B} \sim \pi_t\}$, $\mathcal{D}_{mix} \gets \mathcal{D}_t \cup \{\xi_{1:B} \sim \mathcal{D}_E\}$
\STATE $M_{t+1} \gets M_t - \eta \nabla_M \mathbb{E}_{\mathcal{D}_{mix}}[ - \log(M(s'\mid s, a))]$
\STATE \algcommentlight{Update policy via MBRL w/ resets}
\STATE $\pi_{t+1} \gets \argmax_{\pi \in \Pi} \mathbb{E}_{\substack{h \sim \mathsf{Unif}[1, H] \\ s_h \sim \mathcal{D}_E[h]}} \left[ V^{\pi} \left(s_h \mid -f_t, M_t\right) \right]$
\ENDFOR
\STATE {\bfseries Return } best of $\pi_{1:T}$ on validation data.
\end{algorithmic}
\caption{\textbf{Hy}brid \textbf{P}olicy \textbf{E}mulation w/ \textbf{R}esets (\texttt{HyPER}) \label{alg:hyper}}
\end{algorithm}
Intuitively, this bound tells us that with sufficient inner loop ($M$) and outer loop ($T$) iterations, we can guarantee that we will find a policy with similar performance to that of the expert under the ground-truth reward function. More precisely, it tells us that we need to perform  $O(H^2)$ inner loop steps to avoid compounding errors, assuming realizability of the expert policy. %
Critically, unlike FILTER, \texttt{HyPE} does not require generative model access to the environment. \footnote{An open question for future work is the robustness of 
the hybrid approach to compounding errors when the expert policy \textit{isn't} realizable; existing interactive algorithms like MaxEntIRL \cite{ziebart2008maximum, swamy2021moments} and DAgger \cite{ross2011reduction} are known to be robust to such mis-specification, and we conjecture the same for the hybrid approach espoused here.}

\subsection{\texttt{HyPER}: Model-Based Hybrid Inverse RL}
We now consider how best to design a \emph{model-based} hybrid IRL algorithm, again first considering the forward problem. A common recipe for hybrid model-based RL is to \textit{(1)} fit a model on a mixture of learner and expert data, \textit{(2)} compute the optimal policy in this model, and \textit{(3)} go back to Step \textit{(1)} \citep{ross2012agnostic}. While Step \textit{(2)} doesn't require any real-world interaction, it can still involve an amount of computation in the model that scales with $\exp(H)$. To deal with this concern, \citet{vemula2023virtues} suggest running the No Regret Policy Iteration (NRPI) algorithm of \citet{Ross2014ReinforcementAI} \textit{inside} the model, which comes with strong $\mathsf{poly}(H)$ interaction complexity guarantees. Practically, this looks like model-based RL but with resets to states from the expert's state distribution -- as we've fit this model ourselves, we clearly have generative model access to it to reset. 

\texttt{HyPER} (\textbf{Hy}brid \textbf{P}olicy \textbf{E}mulation with \textbf{R}esets) lift this idea to the space of imitation learning as outlined in Algorithm~\ref{alg:hyper}. In each iteration, \texttt{HyPER} picks both an adversarial reward function and a model, and updates the policy using model-based RL with resets. \texttt{HyPER} can be thought of as running the LAMPS algorithm of \citet{vemula2023virtues} with adversarially chosen rewards, or, equivalently, as running the FILTER algorithm of \citet{Swamy2023InverseRL} inside a model fit in a hybrid fashion. We now prove that the LAMPS algorithm of \citet{vemula2023virtues} satisfies the $\mathsf{ERROr}$ property. %
\begin{lemma} Given any sequence of reward functions $f_{1:T}$, LAMPS picks a sequence of policies $\pi_{1:T}$ that satisfies
\[ \frac{1}{T} \sum_t^T \left(J(\pi_E,  f_t) - J(\pi_t, f_t)\right) \leq (\bar{\epsilon}_{\pi}  + 2 \sqrt{\bar{\epsilon}_M})H^2\]
where $\bar{\epsilon}_{\pi}$ and $\bar{\epsilon}_M$ are the average regret of the policy and model selection subroutines. \hyperref[pf:lem:lamps]{[Proof]}
\label{lem:lamps}
\end{lemma}

As was the case for \texttt{HyPE}, LAMPS satisfying the $\mathsf{ERRoR}$ property directly implies a performance bound for \texttt{HyPER}.

\begin{corollary}[\texttt{HyPER} Performance Bound]
Consider running \texttt{HyPER} (Algorithm \ref{alg:hyper}) for $T$ iterations. Then, we have the following performance guarantee for average policy $\bar{\pi}$:
    \begin{equation}
        J(\pi_E, r) - J(\bar{\pi}, r) \leq (\bar{\epsilon}_{\pi} + 2 \sqrt{\bar{\epsilon}_M}) H^2 + \frac{\mathsf{Reg}_f(T)}{T} H 
    \end{equation}
\end{corollary}
\begin{pf} Follows directly from Theorem \ref{thm:hyirl} and Lemma \ref{lem:lamps}. 
\end{pf}

Observe that this algorithm allows us to gain the computational efficiency benefits of expert resets without needing generative model access to the environment like FILTER. Compared to \texttt{HyPE}, we have to pay $O(H^2)$ in terms of our model learning error. However, this comes with the benefit of only needing to perform policy evaluation rather than policy search in the real world. Thus, we would expect that for problems where we can accurately model the dynamics, \texttt{HyPER} would be more interaction efficient than \texttt{HyPE}.

\subsection{Efficient IRL Battle Royale}
\label{sec:battle}
How does one choose from different flavors of efficient IRL algorithms? The best choice depends on the degree of environment access, type of demonstrations, and whether there are multiple tasks to be performed in a single environment. 

If we assume generative model access to the environment, then FILTER~\cite{Swamy2023InverseRL} seems like the best approach to employ. It comes with strong $\mathsf{poly}(H)$ guarantees on interaction complexity and can also handle scenarios where the demonstrations do not have action labels. However, for many real-world applications such as household robotics, resets to arbitrary states are unrealistic. 

If we assume we can model the environment well, then \texttt{HyPER} mitigates the need for expensive, potentially unsafe exploration in the real world. A learned model also allows for resetting, thus providing strong computational efficiency guarantees similar to FILTER. \texttt{HyPER} is particularly useful in multi-task settings where tasks may differ in terms of reward but share common dynamics, e.g. a home robot solving multiple tasks that all share a common physical setup like a kitchen, as explored in \citet{kim2023learning}.

\texttt{HyPE} requires neither of these assumptions, rendering it the most broadly applicable of the efficient IRL algorithms. However, its interaction efficiency guarantees are strongly tied to the underlying hybrid RL algorithm. For example, to argue that HyQ is more efficient than off-the-shelf fitted Q iteration, one needs strong assumptions like Bellman Completeness of $\mathcal{F}_Q$ and low Bellman Rank \citep{song2022hybrid}. %
Thus, one needs to be base their selection of policy search procedure on the precise characteristics of the problem they are attempting to solve to be assured of efficiency.

We note that \texttt{HyPE} / \texttt{HyPER} are complimentary to FILTER and can be applied in combination to further boost performance, as demonstrated in our antmaze experiments. 

Outside of these, there are other techniques to boost efficiency in IRL. A simple approach is to KL-regularize the learner to a behavior cloning policy \citep{tiapkin2023regularized}, which has proven empirically successful in a variety of problems. However, there are settings where KL regularization leads to undesirable behavior, for example, averaging across differing behavioral modes in the demonstrations. On problems where it is beneficial however, BC regularization can be combined with any of the strategies for improved sample efficiency. Other approaches involve bypassing the need for a reward model altogether and instead using the difference of Q values to implicitly represent rewards \citep{Dvijotham2010InverseOC, garg2021iqlearn} Unfortunately, since $Q$ values depends on the dynamics of the environment, if the dynamics were to change across demonstrations, such approaches may fail to recover consistent estimates of the expert's policy. However, for certain problems where the dynamics are always the same (e.g. language modeling), such approaches can perform well \citep{cundy2023sequencematch}.

\section{Experiments}

\begin{figure*}[t!]
    \centering
    \includegraphics[scale=0.44]{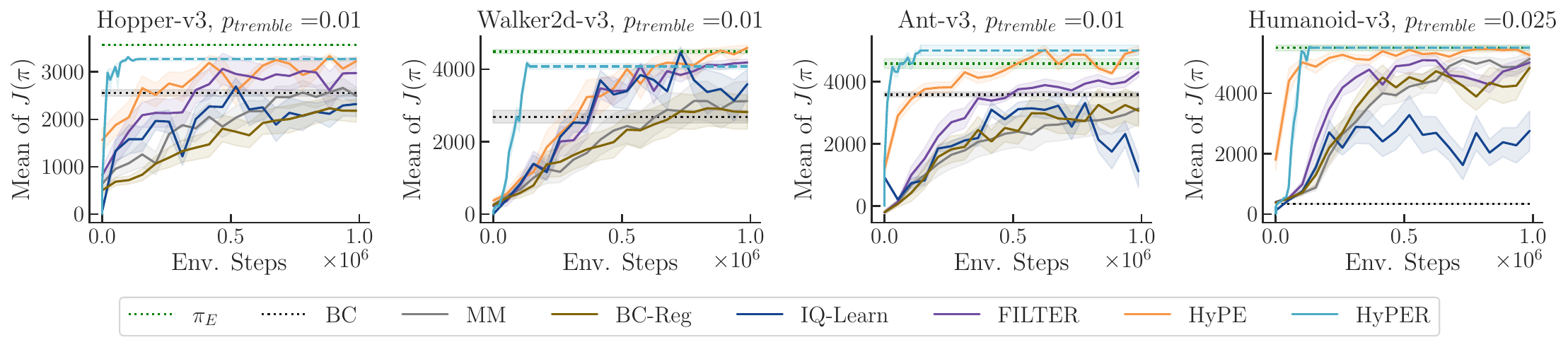}
    \caption{We see \texttt{HyPER} and \texttt{HyPE} achieve the highest reward on the MuJoCo locomotion benchmark. Further, the performance gap increases with the difficulty of the environment (i.e. how far right a plot is in the above figure). We run all model-free algorithms for 1 million environment steps. Due to the higher interaction efficiency of model-based approaches, we only run \texttt{HyPER} for 150k environment steps, after which the last reward is extended horizontally across. We compute standard error across 5 seeds for \texttt{HyPER}, and across 10 seeds for all other algorithms. \UrlBigBreaks{}}
    \label{fig:mujoco-exps}
\end{figure*}

In this section, we aim to answer the following questions:

\begin{enumerate}
    \item\textbf{Are \texttt{HyPE} and \texttt{HyPER} more sample efficient than prior IRL methods?} Since \texttt{HyPE} and \texttt{HyPER} see expert data during their updates, we expect them to converge to expert performance with fewer environment interactions than a standard IRL approaches.
    \item\textbf{Do \texttt{HyPE} and \texttt{HyPER} handle environments with hard exploration challenges better than prior IRL methods?} We conduct experiments on $\texttt{antmaze-large}$, where the learner must control a four-legged agent to navigate to a goal within a maze.
    \item\textbf{Are there cases where a model-based approach provides performance or efficiency gains?} In \texttt{HyPER}, environment interaction is used only for policy evaluation rather than policy search. Thus, we expect \texttt{HyPER} to be even more sample efficient than \texttt{HyPE}. 
\end{enumerate}

We implement \texttt{HyPE} by updating the policy and critic networks in Soft Actor Critic~\citep{Haarnoja2018SoftAO} with expert and learner samples. We implement \texttt{HyPER} by running model-based policy optimization \citep{janner2019trust} and resetting to expert states in the learned model. No reward information is provided in either case, so we also train a discriminator network. We compare against five baselines: BC (behavioral cloning, \citet{pomerleau1988alvinn}), IQLearn~\citep{garg2021iqlearn}, MM (a baseline moment-matching algorithm that uses an integral probability metric instead of Jensen-Shannon divergence as suggested by~\citet{swamy2022minimax}), BC-Reg (MM with an added Mean-Square Error Loss on the actor update), and FILTER (IRL with resets to expert states, ~\citet{Swamy2023InverseRL}). Appendix~\ref{app:impl-details} includes additional implementation details and hyperparameters. We release the code we used for all of our experiments at \textbf{\texttt{\url{https://github.com/jren03/garage}}}.

\noindent\textbf{MuJoCo Experiments.}
On the MuJoCo locomotion benchmark environments~\citep{brockmangym}, all learners are provided with 64 demonstration trajectories \footnote{We perform an ablation of all methods considered in the limited demonstration regime in Appendix \ref{app:finite_sample}.} from an RL expert to mitigate finite sample concerns. Given simple behavior cloning can match expert performance under these conditions \citep{swamy2021moments}, we harden the problem, noise is injected into the environment in the following manner: with probability $p_{tremble}$, a random action is executed instead of the one chosen by the policy. As seen in Figure~\ref{fig:mujoco-exps}, \texttt{HyPE} and \texttt{HyPER} converge much quicker to expert performance compared to baselines, and the gap increases as environments get more difficult (further right in the figure). While our algorithms have the same worst-case performance bound as FILTER, \texttt{HyPE} and \texttt{HyPER} show to be empirically more sample efficient. We hypothesize this is because \texttt{HyPE} and \texttt{HyPER} use expert states and actions, while FILTER only uses expert states. We find these results to be particularly exciting, as \texttt{HyPE} and \texttt{HyPER} algorithms show competitive performance and sample efficiency without needing expert resets. Finally, IQLearn fails to match BC performance on some environments, and shows signs of unstable training on others. We suspect this is due to the need to model environment stochasticity implicitly.

\begin{figure}[h]
    \centering
    \includegraphics[scale=0.4]{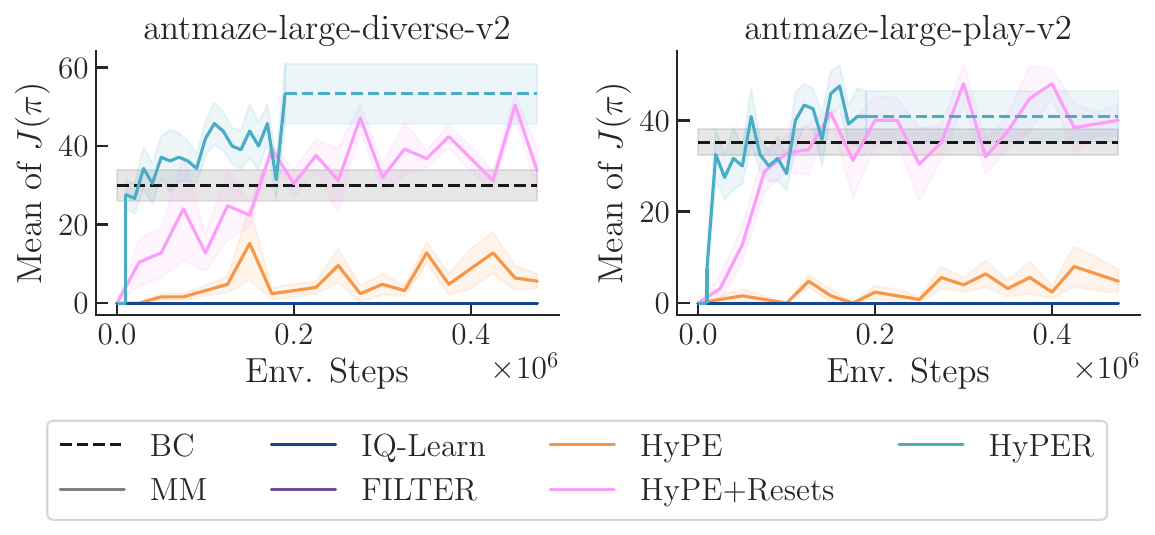}
    \caption{\label{fig:maze-exps}Results on D4RL \texttt{antmaze-large} environment. All interactive baselines achieve 0 reward. While \texttt{HyPE} outperforms prior interactive methods, it does require resets in the environment to beat BC. \texttt{HyPER} is able to surpass BC \emph{without} needing to reset to expert states and match BC performance with roughly 1/10\textit{th} the amount of environment interaction that \texttt{HyPE} + Resets requires. Standard errors are reported across 5 seeds for all algorithms.}
\end{figure}

\noindent\textbf{D4RL Experiments.} Our next set of experiments consider the D4RL~\citep{fu2020d4rl} $\texttt{antmaze-large}$ environments, which is challenging for interactive algorithms that don't use any reward information. We use the standard D4RL dataset and use TD3+BC~\citep{fu2018learning} as our policy optimizer. We use a PSDP-inspired reset strategy for \texttt{HyPER}, where if $T$ is total training steps and $H$ the task horizon, then for each iteration $t$ we reset to the set of expert states falling within a sliding window of size $\kappa \in [0, 1]$:
\begin{equation}
    \left[H \cdot \left(1 - \frac{t}{T}\right), H \cdot \min\left(1, 1 - \frac{t}{T} + \kappa\right) \right].
\end{equation}

In Figure~\ref{fig:maze-exps}, we see that all baseline interactive algorithms (including FILTER \footnote{This does not conflict with the results of \citet{Swamy2023InverseRL}. While they do not explicitly mention it in their paper, they use hybrid training for their strongest results on these environments.}) achieve zero reward. This underscores the important of leveraging expert actions for these environments. While \texttt{HyPE} gets off the ground, it fails to match BC performance. If we combine \texttt{HyPE} with resets in the real environment, we find that we are able to improve past BC performance. 
However, \texttt{HyPER} not only matches the performance of \texttt{HyPE} + Resets with less online interaction, it does so \emph{without needing to do resets in the real environment}. Instead, we perform expert resets within our learned world model.  
To our knowledge, this is the highest performance achieved by an inverse reinforcement learning algorithm on \texttt{antmaze}, including those that require generative model access to the environment. Thus, we believe that \texttt{HyPER} may have interesting applications to real-world robotics tasks.

\section{Conclusion}
Despite the many benefits interactive approaches to imitation learning like inverse RL provide, they suffer from a high level of interaction complexity due to their reduction to \textit{repeated} RL. In response, we derive a new paradigm of efficient IRL algorithms via a novel reduction to \textit{expert-competitive} RL. We then instantiate this reduction with \textit{hybrid}, deriving both model-based and model-free algorithms, and show that we can dramatically reduce the amount of interaction required to compute a strong policy by utilizing expert demonstrations during policy search. By preserving the benefits of interaction in imitation while reducing unnecessary exploration, we believe that we have provided the proper algorithmic foundations to take advantage of recent developments in hybrid RL and model-based RL on problems of interest in domains like robotics.

\section*{Acknowledgments}
JR and SC are supported by the NSF RI \#2312956. ZSW is supported in part by the NSF FAI Award \#1939606, a Google Faculty Research Award, a J.P. Morgan Faculty Award, a Facebook Research Award, an Okawa Foundation Research Grant, and a Mozilla Research Grant. GS would like to thank Yuda Song for many helpful edits and being a hybrid RL oracle throughout the course of this project and Ken Nakamura for comments on proof style.

\section*{Impact Statement}
This paper presents work whose goal is to advance the field of Machine Learning. There are many potential societal consequences of our work, none which we feel must be specifically highlighted here.

\section*{Contribution Statements}
\begin{itemize}
\item \textbf{JR} implemented both algorithms and baselines, performed all experiments and ablations, and released a high-quality open-source implementation.
\item \textbf{GS} proposed the project, derived the key novel reduction, came up with the algorithms, proved all theorems, wrote most of the paper, and provided extensive assistance with debugging all of the experiments.
\item \textbf{SC} came up with the proof of Lemma 3.5 and wrote Section 3.5 of the paper.
\item \textbf{ZSW, JAB,} and \textbf{SC} advised the project. 
\end{itemize}

\bibliography{example_paper}
\bibliographystyle{icml2024}

\newpage
\appendix
\onecolumn
\section{Proofs}
\subsection{Miscellaneous Lemmas}
We begin by proving several lemmas that will be helpful in the following proofs.
\begin{lemma}[Policy Evaluation Lemma, \citep{xie2020q}] For any policy $\pi$ and state-action function $Q$, we have that
    \begin{equation}
        \mathbb{E}_{s_0 \sim \rho}[\mathbb{E}_{a \sim \pi(s_0)}[Q(s, a)]] - J(\pi) = \mathbb{E}_{\xi \sim \pi}\left[\sum_h^H (Q - \mathcal{T}^{\pi}_r Q)(s_h, a_h)\right],
    \end{equation}
    \label{lem:pe}
    where $\mathcal{T}^{\pi}_r$ is the Bellman operator under $\pi$ and $r$.
\end{lemma}
\begin{proof}
    \citet{xie2020q} consider the infinite horizon, discounted setting while we consider the finite horizon, un-discounted setting so we require a slightly different proof.
    \begin{align*}
        \mathbb{E}_{s_0 \sim \rho}[\mathbb{E}_{a \sim \pi(s_0)}[Q(s, a)]] - J(\pi) &= \mathbb{E}_{s_0 \sim \rho}[\mathbb{E}_{a \sim \pi(s_0)}[Q(s, a)]] - \mathbb{E}_{\xi \sim \pi}\left[\sum_{h=0}^H r(s_h, a_h) \right] \\
        &= \mathbb{E}_{s_0 \sim \rho}[\mathbb{E}_{a \sim \pi(s_0)}[Q(s, a)]] - \mathbb{E}_{\xi \sim \pi}\left[\sum_{h=0}^H r(s_h, a_h) \right] \\
        &+ \left(\mathbb{E}_{\xi \sim \pi}\left[\sum_{h=1}^H Q(s_h, a_h) -Q(s_h, a_h) \right]\right) \\
        &= \mathbb{E}_{\xi \sim \pi}\left[\sum_{h=0}^H Q(s_h, a_h) -(r(s_h, a_h) + Q(s_{h+1}, a_{h+1})) \right] \\
        &= \mathbb{E}_{\xi \sim \pi}\left[\sum_{h=0}^H (Q - \mathcal{T}^{\pi}_r Q)(s_h, a_h) \right],
    \end{align*}
    where we define $Q(s_{H+1}, a_{H+1}) = 0$.
\end{proof}

\begin{lemma}[Performance Difference via Advantage in Model (PDAM, \cite{vemula2023virtues})] Given two policies $\pi_E$, $\pi$ and a model $\hat{M}$, we have that
\begin{align}
    J_{M^{\star}}(\pi_E, r) - J_{M^{\star}}(\pi, r) &= \mathbb{E}_{\xi \sim \pi_E, M^{\star}} \left[ \sum_h^H \mathbb{E}_{a \sim \pi_E(s_h)}[\hat{Q}_r(s_h, a)] - \mathbb{E}_{a \sim \pi(s_h)}[\hat{Q}_r(s_h, a)] \right] \nonumber \\ 
    &+ \mathbb{E}_{\xi \sim \pi_E, M^{\star}} \left[ \sum_h^H \mathbb{E}_{\substack{s^{\star}_{h+1} \sim M^{\star} \\ a \sim \pi(s_{h+1}^{\star})}}[\hat{Q}_r(s^{\star}_{h+1}, a)] - 
    \mathbb{E}_{\substack{\hat{s}_{h+1} \sim \hat{M} \\ a \sim \pi(\hat{s}_{h+1})}}[\hat{Q}_r(\hat{s}_{h+1}, a)] \right] 
    \nonumber \\
    &+ \mathbb{E}_{\xi \sim \pi, M^{\star}} \left[ \sum_h^H \mathbb{E}_{\substack{\hat{s}_{h+1} \sim \hat{M} \\ a \sim \pi(\hat{s}_{h+1})}}[\hat{Q}_r(\hat{s}_{h+1}, a)] - \mathbb{E}_{\substack{s^{\star}_{h+1} \sim M^{\star} \\ a \sim \pi(s_{h+1}^{\star})}}[\hat{Q}_r(s^{\star}_{h+1}, a)]  \right]. 
    \nonumber
\end{align}
\end{lemma}
\begin{proof}
    We significantly simply the proof of \cite{vemula2023virtues}. We first break up the performance difference into a sum of three terms.
    \begin{align}
        J_{M^{\star}}(\pi_E, r) - J_{M^{\star}}(\pi, r) &= \left(J_{M^{\star}}(\pi_E, r) - \mathbb{E}_{s_0 \sim \rho}[\mathbb{E}_{a \sim \pi_E(s_0)}[\hat{Q}_r(s_0, a)]] \right) \tag{T1} \\
        &- \left(J_{M^{\star}}(\pi, r) - \mathbb{E}_{s_0 \sim \rho}[\mathbb{E}_{a \sim \pi(s_0)}[\hat{Q}_r(s_0, a)]] \right) \tag{T2} \\ 
        &+ \left(\mathbb{E}_{s_0 \sim \rho}[\mathbb{E}_{a \sim \pi_E(s_0)}[\hat{Q}_r(s_0, a)]] - \mathbb{E}_{s_0 \sim \rho}[\mathbb{E}_{a \sim \pi(s_0)}[\hat{Q}_r(s_0, a)]] \right) \tag{T3}
    \end{align}
We consider each term separately. First, by Lemma \ref{lem:pe}, we have that
\begin{align*}
    \text{(T1)} &= \mathbb{E}_{\xi \sim \pi_E, M^{\star}}\left[ \sum_{h=0}^H (\mathcal{T}^E_r \hat{Q}_r - \hat{Q}_r)(s_h, a_h) \right] \\
    &= \mathbb{E}_{\xi \sim \pi_E, M^{\star}}\left[ \sum_{h=0}^H (\cancel{r(s_h, a_h)} + \mathbb{E}_{\substack{s^{\star}_{h+1} \sim M^{\star}(s_h, a_h) \\ a \sim \pi_E(s^{\star}_{h+1})}}[\hat{Q}_r(s^{\star}_{h+1}, a)]) - (\cancel{r(s_h, a_h)} + \mathbb{E}_{\substack{\hat{s}_{h+1} \sim \hat{M}(s_h, a_h) \\ a \sim \pi_E(\hat{s}_{h+1})}}[\hat{Q}_r(\hat{s}_{h+1}, a)]) \right] \\
    &= \mathbb{E}_{\xi \sim \pi_E, M^{\star}} \left[ \sum_{h=0}^H \left(\mathbb{E}_{\substack{s^{\star}_{h+1} \sim M^{\star}(s_h, a_h) \\ a \sim \pi_E(s^{\star}_{h+1})}}[\hat{Q}_r(s^{\star}_{h+1}, a)] - \mathbb{E}_{\substack{s^{\star}_{h+1} \sim M^{\star}(s_h, a_h) \\ a \sim \pi(s^{\star}_{h+1})}}[\hat{Q}_r(s^{\star}_{h+1}, a)] \right) \right. \\  
    & \left. -\sum_{h=0}^H \left( \mathbb{E}_{\substack{\hat{s}_{h+1} \sim \hat{M}(s_h, a_h) \\ a \sim \pi_E(\hat{s}_{h+1})}}[\hat{Q}_r(\hat{s}_{h+1}, a)] - \mathbb{E}_{\substack{s^{\star}_{h+1} \sim M^{\star}(s_h, a_h) \\ a \sim \pi(s^{\star}_{h+1})}}[\hat{Q}_r(s^{\star}_{h+1}, a)] \right) \right]. \\
\end{align*}
Adding in (T3) to this expression gives us
\begin{align*}
    \text{(T1) + (T3)} &= \mathbb{E}_{\xi \sim \pi_E, M^{\star}} \left[ \sum_{h=0}^H \left(\mathbb{E}_{\substack{a \sim \pi_E(s_h)}}[\hat{Q}_r(s^{\star}_{h+1}, a)] - \mathbb{E}_{\substack{ a \sim \pi(s_h)}}[\hat{Q}_r(s^{\star}_{h+1}, a)]\right) \right. \\
    &\left. +\sum_{h=0}^H \left( \mathbb{E}_{\substack{s^{\star}_{h+1} \sim M^{\star}(s_h, a_h) \\ a \sim \pi(s^{\star}_{h+1})}}[\hat{Q}_r(s^{\star}_{h+1}, a)] - \mathbb{E}_{\substack{\hat{s}_{h+1} \sim \hat{M}(s_h, a_h) \\ a \sim \pi_E(\hat{s}_{h+1})}}[\hat{Q}_r(\hat{s}_{h+1}, a)] \right) \right]. \\
\end{align*}
Next, again by Lemma \ref{lem:pe}, we have
\begin{align*}
    \text{(T2)} &= \mathbb{E}_{\xi \sim \pi, M^{\star}}\left[ \sum_{h=0}^H (\hat{Q}_r - \mathcal{T}^{\pi}_r \hat{Q}_r)(s_h, a_h) \right] \\
    &= \mathbb{E}_{\xi \sim \pi, M^{\star}}\left[ \sum_{h=0}^H  (\cancel{r(s_h, a_h)} + \mathbb{E}_{\substack{\hat{s}_{h+1} \sim \hat{M}(s_h, a_h) \\ a \sim \pi(\hat{s}_{h+1})}}[\hat{Q}_r(\hat{s}_{h+1}, a)]) -
    (\cancel{r(s_h, a_h)} + \mathbb{E}_{\substack{s^{\star}_{h+1} \sim M^{\star}(s_h, a_h) \\ a \sim \pi(s^{\star}_{h+1})}}[\hat{Q}_r(s^{\star}_{h+1}, a)])\right] \\
    &= \mathbb{E}_{\xi \sim \pi, M^{\star}}\left[ \sum_{h=0}^H  ( \mathbb{E}_{\substack{\hat{s}_{h+1} \sim \hat{M}(s_h, a_h) \\ a \sim \pi(\hat{s}_{h+1})}}[\hat{Q}_r(\hat{s}_{h+1}, a)]) -
    ( \mathbb{E}_{\substack{s^{\star}_{h+1} \sim M^{\star}(s_h, a_h) \\ a \sim \pi(s^{\star}_{h+1})}}[\hat{Q}_r(s^{\star}_{h+1}, a)])\right]. \\
\end{align*}
Adding together the preceding results gives the desired bound.
\end{proof}

A direct corollary of this lemma via applying Hölder's inequality to the last two terms is as follows.
\begin{corollary} Define $\tilde{\pi}$ as the trajectory-level average of $\pi$ and $\pi_E$. Then, we have that
\begin{align}
    J_{M^{\star}}(\pi_E, r) - J_{M^{\star}}(\pi, r) &\leq \mathbb{E}_{\xi \sim \pi_E, M^{\star}} \left[ \sum_h^H \mathbb{E}_{a \sim \pi_E(s_h)}[\hat{Q}_r(s_h, a)] - \mathbb{E}_{a \sim \pi(s_h)}[\hat{Q}_r(s_h, a)] \right] \nonumber \\
    &+ 2 H^2 \mathbb{E}_{s, a \sim d_{\tilde{\pi}}}\left[D_{\text{TV}}(M^{\star}(s, a), \hat{M}(s, a)) \right], \nonumber
\end{align}
where $D_{TV}$ denotes the total variation distance between two distributions.
\label{corr:pdam}
\end{corollary}

\subsection{Proof of Lemma \ref{lem:lamps}}
\label{pf:lem:lamps}
\begin{proof}
     We define two losses, one for each player:
    \begin{equation}
        \ell_{t+1}(M) = \mathbb{E}_{s,a \sim \Tilde{\pi}_i}\left[D_{KL}(M^{\star}(s, a) || \hat{M}(s, a)) \right]
    \end{equation}
    \begin{equation}
        \ell_{t+1}(\pi) = \frac{1}{H^2} \left(\mathbb{E}_{\xi \sim \pi_E, M^{\star}} \left[\sum_h^H \mathbb{E}_{a \sim \pi(s_h)} \left [ Q^{\pi_t}_{M_{t+1}, f_{t+1}}(s_h, a)\right] \right] \right)
    \end{equation}
Observe that minimizing the first involves an online convex optimization oracle over $\mathcal{M}$, and the second an online cost-sensitive classification oracle over $\Pi$. Also observe that, when summed, the policy and model losses bound the PDAM. To satisfy the $\mathsf{ERRoR}$ property, we need to be able to upper bound
\begin{equation}
    \frac{1}{H} \sum_t^T J_{M^{\star}}(\pi_E, f_t) - J_{M^{\star}}(\pi_t, f_t).
\end{equation}

Applying Corollary \ref{corr:pdam} to the $t$th term in the summation, we get 
\begin{align*}
    J_{M^{\star}}(\pi_E, f_t) - J_{M^{\star}}(\pi_i, f_t) &\leq \mathbb{E}_{\xi \sim \pi_E, M^{\star}} \left[ \sum_h^H \mathbb{E}_{a \sim \pi_E(s_h)}[\hat{Q}_{f_t}(s_h, a)] - \mathbb{E}_{a \sim \pi(s_h)}[\hat{Q}_{f_t}(s_h, a)] \right] \nonumber \\
    &+ 2 H^2 \mathbb{E}_{s, a \sim d_{\tilde{\pi}_{t-1}}}\left[D_{\text{TV}}(M^{\star}(s, a), M_t(s, a)) \right].
\end{align*}
Recall that running NRPI for $K$ iterations guarantees that $\ell_t(\pi_E) - \ell_t(\pi_t) \leq \bar{\epsilon}^K_{\pi}$, where $\pi_t$ is the best of the $K$ policy iterates generated (or their average). This directly implies that
\begin{align*}
    J_{M^{\star}}(\pi_E, f_t) - J_{M^{\star}}(\pi_t, f_t) \leq \bar{\epsilon}_{\pi}^K H^2 + 2 H^2 \mathbb{E}_{s, a \sim d_{\tilde{\pi}_{t-1}}}\left[D_{\text{TV}}(M^{\star}(s, a), M_t(s, a)) \right].
\end{align*}
Via the definition of the regret of our $M$ strategy, we have that
\begin{equation}
    \min_{M \in \mathcal{F}_M} \frac{1}{T} \sum_t^T \ell_t(M_t) - \ell_t(M)  \leq \bar{\epsilon}_M \Rightarrow  \frac{1}{T} \sum_t^T \ell_t(M_t) \leq \bar{\epsilon}_M + \min_{M \in \mathcal{F}_M} \frac{1}{T} \sum_t^T \ell_t(M).
\end{equation}
Clearly, $\ell_i(M^{\star}) = 0$ and recall that $M^{\star} \in \mathcal{M}$. Combining these facts with the non-negativity of the KL Divergence, we have that $\min_{M \in \mathcal{M}} \frac{1}{N} \sum_i^N \ell_i(M) = 0$. We can now apply Pinkser's and Jensen's inequalities to simplify the remaining terms:
\begin{align*}
   &\quad \, \frac{1}{T} \sum_{t=1}^T \mathbb{E}_{s, a \sim d_{\tilde{\pi}_{t-1}}}\left[D_{\text{KL}}(M^{\star}(s, a), M_t(s, a)) \right] \leq \bar{\epsilon}_M \\
&\Rightarrow \frac{1}{T} \sum_{t=1}^T \mathbb{E}_{s, a \sim d_{\tilde{\pi}_{t-1}}}\left[D_{\text{TV}}(M^{\star}(s, a), M_t(s, a))^2 \right] \leq \bar{\epsilon}_M \\
&\Rightarrow \frac{2 H^2}{T} \sum_{t=1}^T \mathbb{E}_{s, a \sim d_{\tilde{\pi}_{t-1}}}\left[D_{\text{TV}}(M^{\star}(s, a), M_t(s, a)) \right] \leq 2H^2 \sqrt{\bar{\epsilon}_M}.
\end{align*}
Collecting terms gives us the desired bound.
\end{proof}

\clearpage
\section{Implementation Details}\label{app:impl-details}

We use Optimistic Adam~\citep{daskalakis2017training} for all policy and discriminator optimization, and gradient penalties~\citep{gulrajani2017improved} to stabilize our discriminator training for all algorithms. Our policies, value functions, and discriminators are all 2-layer ReLu networks with a hidden size of 256. We sample 4 trajectories to use in the discriminator update at the end of each outer-loop iteration, and a batch size of 4096. In all 4 IRL variants (\texttt{HyPE}, \texttt{HyPER}, FILTER, MM), we re-label the data with the current reward function during policy improvement, rather than keeping the labels that were set when the data was added to the replay buffer, which we empirically observe to increase performance. This is different from standard IRL implementations and might be of independent interest.

\subsection{MuJoCo Tasks}

We detail below the specific implementations used for all MuJoCo experiments (Ant, Hopper, Humanoid, Walker). To clearly highlight the differences between our algorithms and the baselines, we enumerate separate sections for each.

\noindent\textbf{Discriminator.}
For our discriminator, we start with a learning rate of $8e-4$ and decay it linearly over outer-loop iterations. For all model-free MuJoCo experiments, we update the discriminator every 10,000 actor steps. For model-based MuJoCo experiments, we update the discriminator every 2,000 actor steps for Hopper, and every 10,000 actor steps for Ant, Humanoid, and Walker.

\noindent\textbf{Baselines}.
For IQLearn~\citep{garg2021iqlearn}, we take the exact hyperparameters released in the original repository, but increase the expert memory size to be the same size as all other algorithms. For MuJoCo tasks, this is set to be 64,000 transition tuples. For MM and FILTER baselines, we follow the exact hyperparameters in~\citet{Swamy2023InverseRL} with a small modification of updating the discriminator every 10,000 actor steps instead of the 5,000 in the original repository. Finally, we train all behavioral cloning baselines for 300k steps for Ant, Hopper, and Humanoid, and 500k steps for Walker2d. For BC-Reg, we add the MSELoss to the usual SAC actor update, weighted by $\lambda$. We do a  sweep over $\lambda = \{0.5, 1.0, 2.5, 5.0, 7.5, 10.0\}$ and take the $\lambda$ that gives the highest average performance. We report these values in Table~\ref{table:bc_reg_lambda} for the four MuJoCo environments used.

\begin{table}[h]
\begin{center}
\begin{small}
\begin{sc}
\setlength{\tabcolsep}{2pt}
\begin{tabular}{lccccccccccr}
\toprule
 Environment & & & & & & $\lambda$ \\
\midrule
Ant-v3 & & & & & & 1.0 \\
Humanoid-v3 & & & & & & 0.5 \\
Hopper-v3 & & & & & & 1.0 \\
Walker2d-v3 & & & & & & 0.5 \\
\bottomrule
\end{tabular}
\end{sc}
\end{small}
\end{center}
\caption{\label{table:bc_reg_lambda} Final $\lambda$ values for BC-Reg baseline.}
\end{table}

\noindent\textbf{HyPE}.
For \texttt{HyPE}, we use the Soft Actor Critic~\citep{Haarnoja2018SoftAO} implementation provided by~\citet{raffin2019stable} with the hyperparameters in Table~\ref{table:stbsln3params}. 

\begin{table}[h]
\begin{center}
\begin{small}
\begin{sc}
\setlength{\tabcolsep}{2pt}
\begin{tabular}{lccccccccccr}
\toprule
 Parameter & Value \\
\midrule
 buffer size & 1e6 \\
 batch size & 256 \\
 $\gamma$ & 0.98 \\
 $\tau$ & 0.02 \\
 Training Freq. & 64 \\
 Gradient Steps & 64 \\
 Learning Rate & Lin. Sched. 7.3e-4 \\
 policy architecture & 256 x 2 \\
 state-dependent exploration & true \\
 training timesteps & 1e6 \\
\bottomrule
\end{tabular}
\end{sc}
\end{small}
\end{center}
\caption{\label{table:stbsln3params} Hyperparameters for \texttt{HyPE} using SAC.}
\end{table}

\noindent\textbf{HyPER.}
For \texttt{HyPER}, we use the implementation from~\citet{pineda2021mbrl} with modifications on the actor update according to~\citet{vemula2023virtues}, and turn on the entropy bonus. The model is provided the same demonstration dataset of 64,000 transition tuples as model-free experiments. We use an ensemble of discriminators equal to the model ensemble size, and take the minimum value of the ensemble each forward pass. Further, we find that clipping the discriminator output, adding the same weight decay to the model and actor networks, and using an exponential moving average of the actor weights during inference helps stabilize performance for some environments. We list the specific hyperparameters used for each environment in~\crefrange{table:ant-params}{table:walker-params}.

\subsection{D4RL Tasks}~\label{app:d4rl-impl}

For the two \texttt{antmaze-lage} tasks, we use the data provided by \citet{fu2020d4rl} as out expert demonstrations. We append goal information to the observation for all algorithms following the example in \citet{Swamy2023InverseRL}. For our policy optimizer in every algorithm other than IQLearn, we build upon the TD3+BC implementation of~\citet{fujimoto2021minimalist} with the default hyperparameters. 

\noindent\textbf{Discriminator.}
For our discriminator, we start with a learning rate of $8e-3$ and decay it linearly over outer-loop iterations. For all model-free Antmaze experiments, we update the discriminator every 5,000 actor steps. For all model-based Antmaze experiments, we update the discriminator every 2,000 actor steps. 

\noindent\textbf{Baselines.} 
For behavioral cloning, we run the TD3+BC optimizer for 500k steps while zeroing out the component of the actor update that depends on rewards. We use $\alpha=1.0$ for FILTER. We provide all baselines with the same data provided to \texttt{HyPE} and \texttt{HyPER} consisting of the entire D4RL dataset for both \texttt{antmaze-large} environments. MM and FILTER are pretrained with 10,000 steps of behavioral cloning on the expert dataset.

\noindent\textbf{HyPE.}
Both \texttt{HyPE} and \texttt{HyPE+Resets} use the same TD3+BC optimizer and hyperparameters for the actor as MM and FILTER, and is pretrained with 10,000 steps of behavioral cloning. \texttt{HyPE+Resets} uses $\alpha=1.0$ to always reset to expert states.

\noindent\textbf{HyPER.}
To stabilize performance for \texttt{HyPER}, we pretrain the model on the offline dataset. In addition to the hyperparameters from previous algorithms, we use the exponential moving average of the actor weights and add a CosineAnnealing decay on both the actor and critic. Within the learned model, we perform $n$-step updates backwards in time as inspired by~\citet{Bagnell2003PolicySB} and~\citet{hester2008dqfd} by resetting to a sliding window of expert states within the learned model. Specifically, if $T$ is the total number of training steps and $H$ the horizon of the environment, then for each iteration $t$ we reset to the set of expert states falling within a sliding window of size $\kappa \in [0, 1]$:
\begin{equation*}
    \left[H \cdot \left(1 - \frac{i}{T}\right), H \cdot \min\left(1, 1 - \frac{i}{T} + \kappa \right) \right].
\end{equation*}

We set $\kappa=0.05$ in practice. Additional details and visualizations can be found in Appendix~\ref{app:maze-pretrain}. Finally, we provide the model-based hyperparameters for both \texttt{antmaze-large} environments in Table~\ref{table:antmaze-params}.

\clearpage
\section{Antmaze Model Pretraining}\label{app:maze-pretrain}

\begin{figure*}[h]
    \centering
    \includegraphics[scale=0.32]{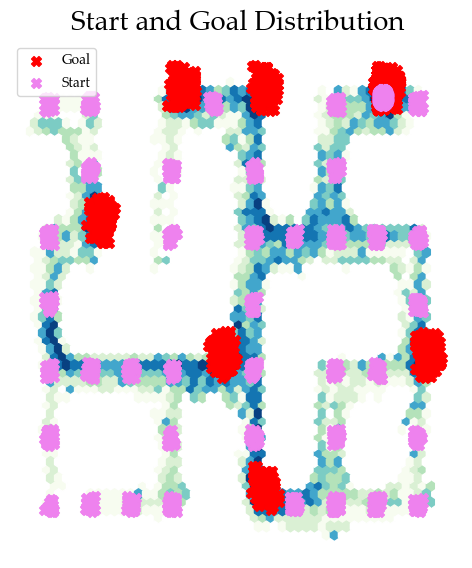}
    \includegraphics[scale=0.32]{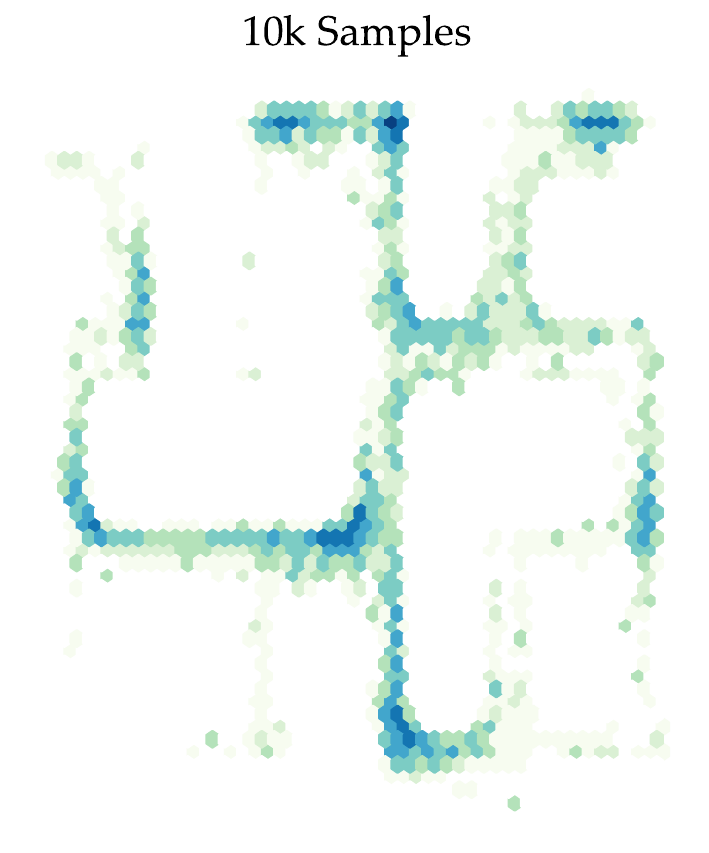}
    \includegraphics[scale=0.32]{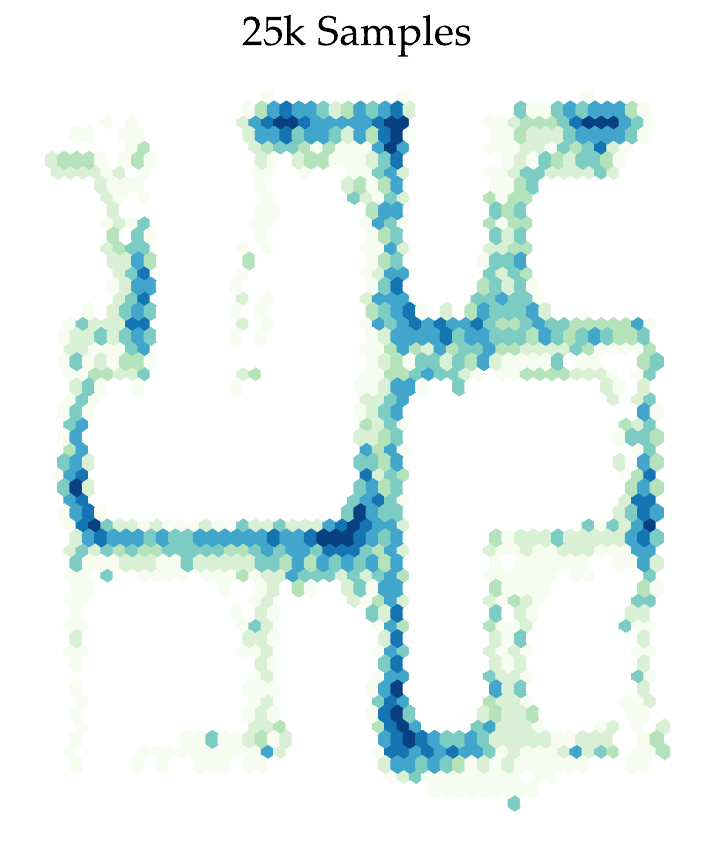}
    \includegraphics[scale=0.32]{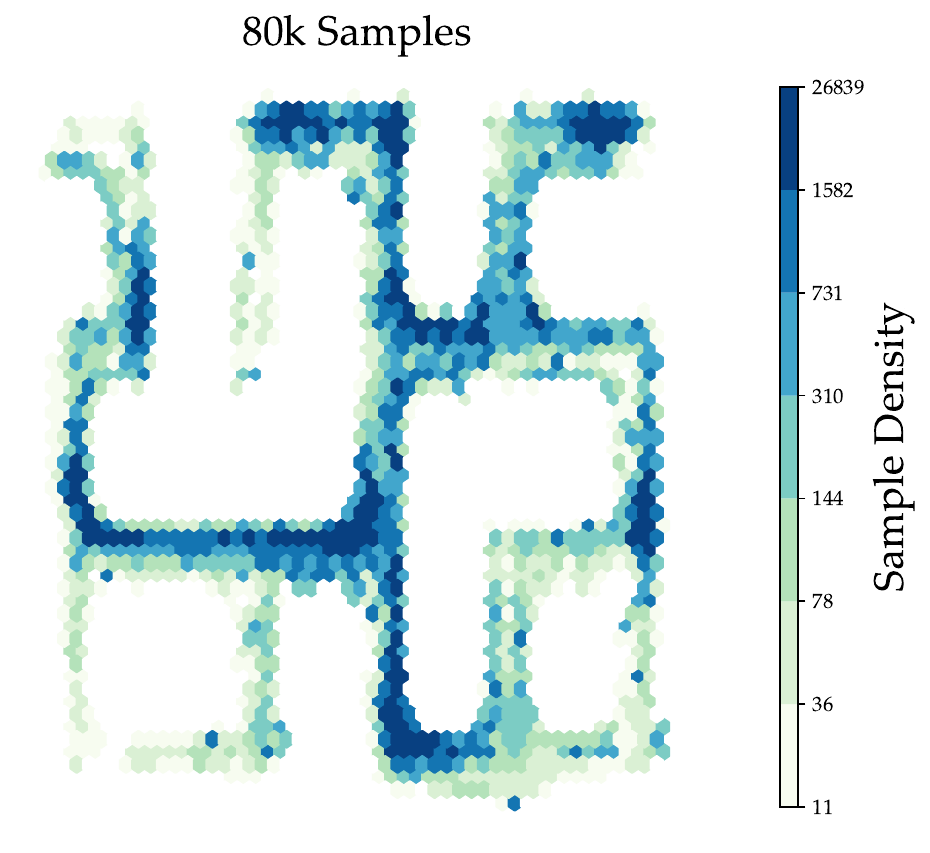}
    \caption{The leftmost plot shows the start and goal distributions of the \texttt{antmaze-large-play} environment. In the subsequent three plots, we show the state visitation frequency across the maze with varying number of samples from the offline dataset. We find a sufficiently large enough sample is necessary to ensure accurate transition tuples in training.}
    \label{fig:maze-pretraining}
\end{figure*}

We visualize below the impact of offline dataset size when pretraining the model for \texttt{antmaze-large}. In the leftmost graph of Figure~\ref{fig:maze-pretraining}, we plot the start and goal distributions for \texttt{antmaze-large-play} in orange and red respectively. In the following three plots, we show the state visitation frequency taking various proportions from the offline data. Notably, with fewer samples, there are regions of the maze with extremely low data coverage, such as the bottom left corner. A model trained on 10k or 25k samples may therefore learn inaccurate dynamics in those regions, leading to unreliable transition dynamics and thus unreliable policy in those regions. Thus by taking 80k samples to pretrain the model and decreasing model update frequency, we ensure that the learner is able to receive sufficiently accurate transition tuples in training. 

\clearpage

\section{Additional Ablations}\label{app:ablations}

\subsection{Classical Adversarial Imitation Learning Methods}

\begin{figure*}[h]
    \centering
    \includegraphics[scale=0.25]{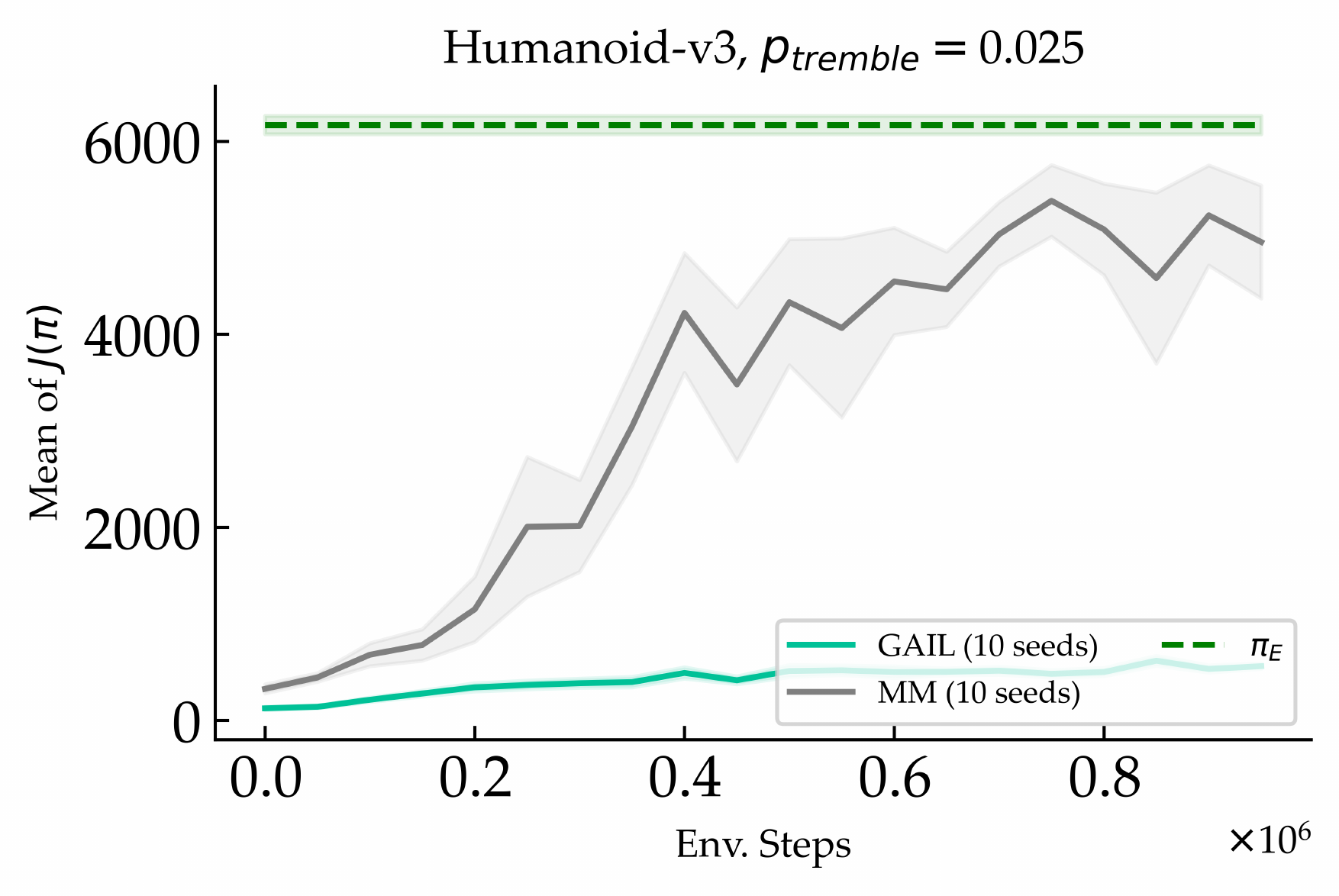}
    \caption{Comparison of our $\texttt{MM}$ baseline against $\texttt{GAIL}$~\citep{ho2016generative}, reporting the average and standard error over 10 seeds. While $\texttt{GAIL}$ make gradual improvement over 1 million environment interactions, it is far from the performance of $\texttt{MM}$.}
    \label{fig:mm_vs_gail}
\end{figure*}

Our inverse RL baseline $\texttt{MM}$ can be thought of as a significantly improved variant of the classical methods such as $\texttt{GAIL}$~\citep{ho2016generative}. More explicitly, we implement the four changes to $\texttt{GAIL}$ suggested by~\citep{swamy2022minimax} which, when combined, lead to a significant improvement over off-the-shelf implementations of $\texttt{GAIL}$. Explicitly, these are 1) using a Wasserstein GAN loss rather than the original GAN loss, 2) using SAC instead of PPO, 3) using gradient penalties in the loss function, and 4) using the Optimistic Adam Optimizer. We would like to point readers to Appendix C of~\citet{swamy2022minimax} for ablations on each of these components. To ablate the joint benefits of these modifications, we compare our $\texttt{MM}$ baseline to $\texttt{GAIL}$ (as implemented in \href{https://github.com/ikostrikov/pytorch-a2c-ppo-acktr-gail}{https://github.com/ikostrikov/pytorch-a2c-ppo-acktr-gail}, a popular implementation with more than 3.5k Github stars) on the Humanoid environment, and report the average performance over 10 seeds. Figure~\ref{fig:mm_vs_gail} shows that while $\texttt{GAIL}$ achieves some improvement over 1 million environment interactions, it massively under performs $\texttt{MM}$, reaffirming the fact that $\texttt{MM}$ is a very strong baseline for standard IRL in and of itself.

\subsection{Low Data Regime}
\label{app:finite_sample}
\begin{figure*}[h]
    \centering
    \includegraphics[scale=0.25]{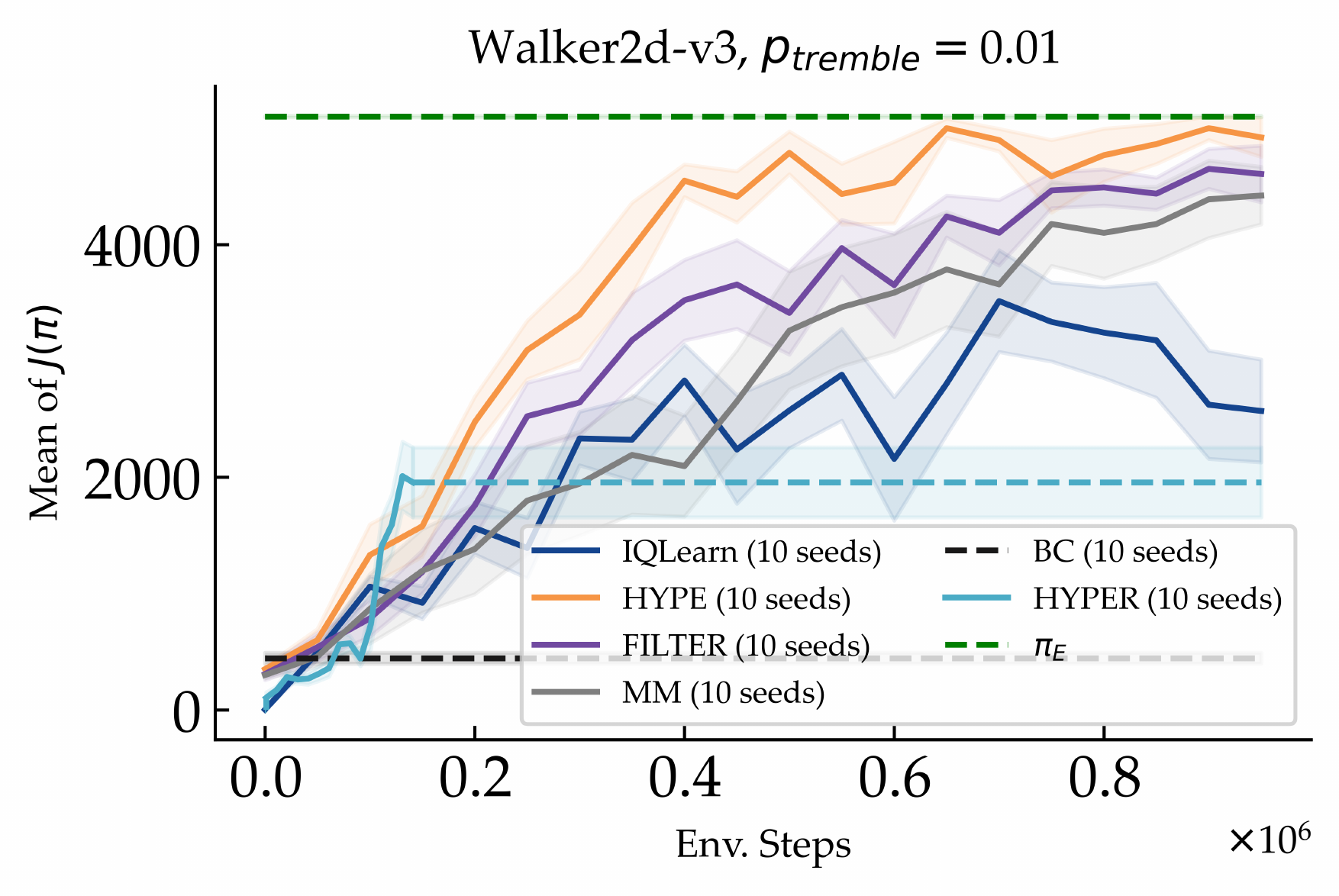}
    \caption{Mean and standard error of various algorithms when provided only 5 expert demonstrations over 10 seeds. $\texttt{HyPE}$ still outperforms existing baselines. $\texttt{HyPER}$ suffers a performance drop, which is expected as low amounts of expert data may not be sufficient for fitting a good model. We believe there are various ways to improve model fitting in practice in the presence of few expert demonstrations, such as multi-task or robot play data.}
    \label{fig:low_data}
\end{figure*}

We present in Figure~\ref{fig:low_data} an ablation of the performance of various algorithms in the low-data regime. Specifically, we take just 5 trajectories to train each algorithm, as opposed to the 64 used in the main experiments. We observe that $\texttt{HyPE}$ still has the quickest convergence over other model-free IRL baselines, even in the low-data regime. $\texttt{HyPER}$ performs rather poorly in the low-data regime, which we hypothesize is due to the difficulty of learning an accurate model from such limited data. However, given a world model can be learned from other data sources which might be more plentiful in practice (e.g. suboptimal demonstrations, multi-task demonstrations, and even robot play data), we believe there are multiple remedies to this issue in practice.

\begin{table}[h]
\begin{center}
\begin{small}
\begin{sc}
\setlength{\tabcolsep}{2pt}
\begin{tabular}{lccccccccccr}
\toprule
 Parameter & Value \\
\midrule
Expert Dataset Size & 64000 \\
Exploration Sample Size & 64000 \\
Model Ensemble Size & 7 \\
Model Ensemble Elite Number & 5 \\
Model Learning Rate & 3e-4 \\
Model Weight Decay & 5e-5 \\
Model Batch Size & 256 \\
Model Train Frequency & 125 \\
Model Hidden Dims Size & 400 \\
Model Clip Output & True \\
Discriminator Clip Output & True \\
Discriminator Weight Decay & True \\
Discriminator Ensemble Size & 7\\
Schedule Model LR & False \\
Schedule Policy LR & False \\ 
EMA policy weights & False \\
Number policy updates per step & 30 \\
Policy updates every steps & 1 \\
Rollout step in learned model & 400 \\
Rollout Length & 1 $\to$ 25 \\
Policy Type & Stochastic Gaussian Policy \\
\bottomrule
\end{tabular}
\end{sc}
\end{small}
\end{center}
\caption{\label{table:ant-params} Hyperparameters for \texttt{Ant-v3}.}
\end{table}

\begin{table}[h]
\begin{center}
\begin{small}
\begin{sc}
\setlength{\tabcolsep}{2pt}
\begin{tabular}{lccccccccccr}
\toprule
 Parameter & Value \\
\midrule
Expert Dataset Size & 64000 \\
Exploration Sample Size & 64000 \\
Model Ensemble Size & 7 \\
Model Ensemble Elite Number & 5 \\
Model Learning Rate & 3e-4 \\
Model Weight Decay & 5e-5 \\
Model Batch Size & 256 \\
Model Train Frequency & 125 \\
Model Hidden Dims Size & 400 \\
Model Clip Output & False \\
Discriminator Clip Output & False \\
Discriminator Weight Decay & False \\
Discriminator Ensemble Size & 7\\
Schedule Model LR & False \\
Schedule Policy LR & False \\ 
EMA policy weights & False \\
Number policy updates per step & 30 \\
Policy updates every steps & 1 \\
Rollout step in learned model & 400 \\
Rollout Length & 1 $\to$ 25 \\
Policy Type & Stochastic Gaussian Policy \\
\bottomrule
\end{tabular}
\end{sc}
\end{small}
\end{center}
\caption{\label{table:hopper-params} Hyperparameters for \texttt{Hopper-v3}.}
\end{table}

\begin{table}[h]
\begin{center}
\begin{small}
\begin{sc}
\setlength{\tabcolsep}{2pt}
\begin{tabular}{lccccccccccr}
\toprule
 Parameter & Value \\
\midrule
Expert Dataset Size & 64000 \\
Exploration Sample Size & 64000 \\
Model Ensemble Size & 7 \\
Model Ensemble Elite Number & 5 \\
Model Learning Rate & 3e-4 \\
Model Weight Decay & 5e-5 \\
Model Batch Size & 256 \\
Model Train Frequency & 125 \\
Model Hidden Dims Size & 400 \\
Model Clip Output & False \\
Discriminator Clip Output & False \\
Discriminator Weight Decay & False \\
Discriminator Ensemble Size & 7\\
Schedule Model LR & False \\
Schedule Policy LR & True \\ 
EMA policy weights & True \\
Number policy updates per step & 20 \\
Policy updates every steps & 2 \\
Rollout step in learned model & 400 \\
Rollout Length & 1 $\to$ 25 \\
Policy Type & Stochastic Gaussian Policy \\
\bottomrule
\end{tabular}
\end{sc}
\end{small}
\end{center}
\caption{\label{table:humanoid-params} Hyperparameters for \texttt{Humanoid-v3}.}
\end{table}

\begin{table}[h]
\begin{center}
\begin{small}
\begin{sc}
\setlength{\tabcolsep}{2pt}
\begin{tabular}{lccccccccccr}
\toprule
 Parameter & Value \\
\midrule
Expert Dataset Size & 64000 \\
Exploration Sample Size & 1000 \\
Model Ensemble Size & 7 \\
Model Ensemble Elite Number & 5 \\
Model Learning Rate & 3e-4 \\
Model Weight Decay & 5e-5 \\
Model Batch Size & 256 \\
Model Train Frequency & 125 \\
Model Hidden Dims Size & 200 \\
Model Clip Output & True \\
Discriminator Clip Output & True \\
Discriminator Weight Decay & True \\
Discriminator Ensemble Size & 7\\
Schedule Model LR & True \\
Schedule Policy LR & True \\ 
EMA policy weights & True \\
Number policy updates per step & 20 \\
Policy updates every steps & 2 \\
Rollout step in learned model & 400 \\
Rollout Length & 1 $\to$ 25 \\
Policy Type & Stochastic Gaussian Policy \\
\bottomrule
\end{tabular}
\end{sc}
\end{small}
\end{center}
\caption{\label{table:walker-params} Hyperparameters for \texttt{Walker-v3}.}
\end{table}

\begin{table}[h]
\begin{center}
\begin{small}
\begin{sc}
\setlength{\tabcolsep}{2pt}
\begin{tabular}{lccccccccccr}
\toprule
 Parameter & Value \\
\midrule
Expert Dataset Size & 999000 \\
Exploration Sample Size & 10000 \\
Model Ensemble Size & 7 \\
Model Ensemble Elite Number & 5 \\
Model Learning Rate & 3e-4 \\
Model Weight Decay & 5e-5 \\
Model Batch Size & 256 \\
Model Train Frequency & 1000 \\
Model Hidden Dims Size & 200 \\
Model Clip Output & False \\
Discriminator Clip Output & False \\
Discriminator Weight Decay & False \\
Discriminator Ensemble Size & 1\\
Schedule Model LR & True \\
Schedule Policy LR & True \\ 
EMA policy weights & True \\
Number policy updates per step & 20 \\
Policy updates every steps & 1 \\
Rollout step in learned model & 400 \\
Rollout Length & 1 $\to$ 25 \\
Policy Type & Stochastic Gaussian Policy \\
\bottomrule
\end{tabular}
\end{sc}
\end{small}
\end{center}
\caption{\label{table:antmaze-params} Hyperparameters for \texttt{antmaze-large}.}
\end{table}

\end{document}